\pdfoutput=1

\documentclass[11pt]{article}

\usepackage{acl}

\usepackage{times}
\usepackage{latexsym}
\usepackage{colortbl}
\usepackage{tabulary}
\usepackage{etoolbox}
\usepackage{amsfonts}
\usepackage{amsmath, amssymb, mathtools}
\usepackage[T1]{fontenc}

\usepackage[utf8]{inputenc}

\usepackage{microtype}

\usepackage{inconsolata}
\usepackage{times}
\usepackage{latexsym}
\usepackage{graphicx}
\usepackage{amsmath}
\usepackage{amssymb}
\usepackage{booktabs}
\usepackage{textcomp}
\usepackage{multirow}
\usepackage{array}
\usepackage{amsfonts}  
\usepackage{subfigure}
\usepackage{enumitem}
\usepackage{color}
\usepackage{xcolor}
\usepackage{hyperref}
\usepackage{bbm}
\usepackage{tabularx, booktabs}
\newcolumntype{Y}{>{\centering\arraybackslash}X}
\newcolumntype{L}{>{\arraybackslash}X}

\usepackage{tablefootnote}
\usepackage{subcaption}

\usepackage[T1]{fontenc}

\usepackage[utf8]{inputenc}

\usepackage{microtype}

\usepackage{xparse}
\definecolor{bluepigment}{rgb}{0.2, 0.2, 0.6}
\definecolor{emerald}{rgb}{0.31, 0.78, 0.47}
\definecolor{orange}{rgb}{0.91, 0.41, 0.17}
\newcommand{\ours}{\textsc{OPAL}}
\newcommand{\taskname}{\textsc{Text2DB}}

\ifx\nocomment\undefined
    \NewDocumentCommand{\heng}
    { mO{} }{\textcolor{red}{\textsuperscript{\textit{Heng}}\textsf{\textbf{\small[#1]}}}}
    \NewDocumentCommand{\jh}
    { mO{} }{\textcolor{purple}{\textsuperscript{\textit{Jiawei}}\textsf{\textbf{\small[#1]}}}}
    \NewDocumentCommand{\zoey}
    { mO{} }{\textcolor{bluepigment}{\textsuperscript{\textit{Zoey}}\textsf{\textbf{\small[#1]}}}}

\else
    \NewDocumentCommand{\heng}
    { mO{} }{\textcolor{red}{}}
    \NewDocumentCommand{\jh}
    { mO{} }{\textcolor{purple}{}}
    \NewDocumentCommand{\zoey}
    { mO{} }{\textcolor{bluepigment}{}}
\fi

\title{
\textsc{Text2DB} : Integration-Aware Information Extraction \\with Large Language Model Agents \\

} 

\author{Yizhu Jiao, Sha Li, Sizhe Zhou, Heng Ji, Jiawei Han  \\
  University of Illinois Urbana-Champaign \\
  \texttt{yizhuj2@illinois.edu} \\}

\begin{document}
\maketitle
\begin{abstract}
The task of information extraction (IE) is to extract structured knowledge from text. 
However, it is often not straightforward to utilize IE output due to the mismatch between the IE ontology 
and the downstream application needs. 
We propose a new formulation of IE \taskname\ that emphasizes the integration of IE output and the target database (or knowledge base). Given a user instruction, a document set, and a database, our task requires the model to update the database with values from the document set to satisfy the user instruction.
This task requires understanding user instructions for \textit{what to extract} and adapting to the given DB/KB schema for \textit{how to extract} on the fly. 
To evaluate this new task, we introduce a new benchmark featuring common demands such as data infilling, row population, and column addition. 
In addition, we propose an LLM agent framework \ours\ (Observe-Plan-Analyze LLM) which includes an Observer component that interacts with the database, the Planner component that generates a code-based plan 
with calls to IE models, and the Analyzer component that provides feedback regarding code quality before execution.
Experiments show that \ours\ can successfully adapt to diverse database schemas by generating different code plans and calling the required IE models. We also highlight difficult cases such as dealing with large databases with complex dependencies and extraction hallucination, which we believe deserve further investigation. 

\end{abstract}

\section{Introduction}

 \begin{figure}
     \centering
     \includegraphics[width=0.5\textwidth]{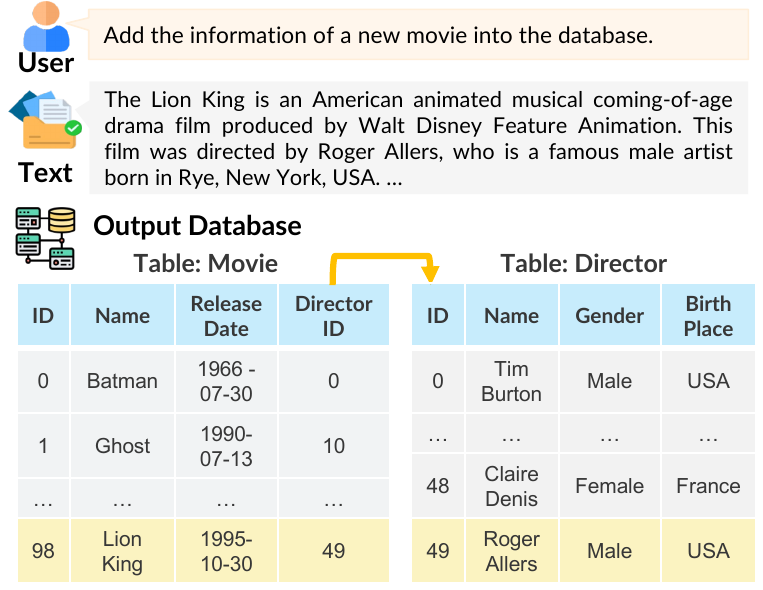}
     \caption{Our \taskname\ task is defined over a database, a user instruction, and a document set.
     The model aims to fulfill the user instruction by updating the database with values (shown in yellow) extracted from text. In this example, the input database has two tables linked with the foreign key constraint (\texttt{DirectorID} in the Movie table refers to \texttt{ID} of the Director table). 
     } 
     \label{fig:intro} 
 \end{figure}

 \begin{figure*}[t]
     \centering
     \includegraphics[width=\textwidth]{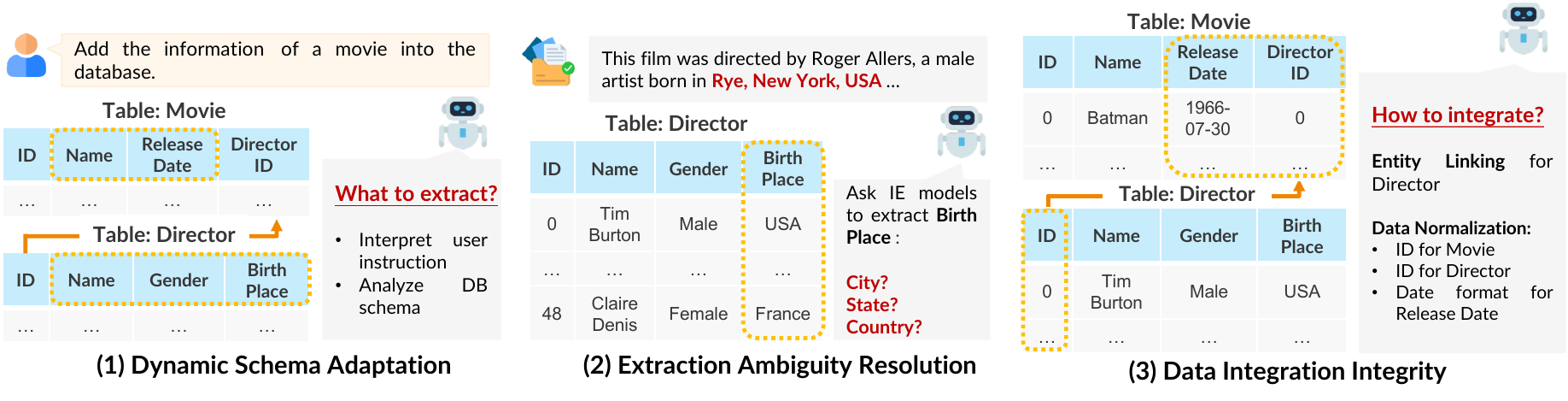}
     \caption{Three major challenges of the \taskname\ task: (1) dynamically decide what to extract by analyzing complex database schemas and interpreting user instructions; (2) resolve extraction ambiguity to ensure extracted values match the semantics and granularity of existing database content; (3) integrate the extracted data into the database while maintaining integrity and consistency.}
     \label{fig:challenge}
 \end{figure*}

Text has always been seen as a rich source of information, and information extraction (IE) is defined as the task of extracting knowledge from unstructured text. However, a long-overlooked question is what counts as ``relevant knowledge'': the entity, relation, and event types that require extraction \cite{ding2021few, wan2023gpt, Li2021doclevel}. Current methods sidestep this question by either assuming that ``relevant knowledge'' is given by the ontology \cite{weischedel2013ontonotes} in the closed domain setting or assuming that all knowledge is relevant in the OpenIE setting \cite{DBLP:conf/dexaw/MuhammadKGCW20}. 
We argue that the scope of relevant knowledge is highly dependent on the downstream task, especially when IE output needs to be ingested into databases or knowledge bases. 
We call such a setting \textit{integration-aware information extraction}, where we take a holistic view and consider both the source of IE and also the consumer of IE results. 
In the database community, integration refers to the alignment between schemas of different databases. We borrow this term to refer to the alignment of IE output and the target database, or in other words, the integration of structured and unstructured information.
Data integration of IE results is critical as (1) a database system provides the infrastructure to support large-scale data management and execution of complex analytical queries; and (2) real-life applications often involve structured data stored in databases that complement IE results (\textit{e.g.,} an e-commerce website has a proprietary product database but might wish to extract user feedback from comments). 

As an instance of this \textit{integration-aware IE} setting, we propose a new task \taskname. Specifically, each instance includes a target database (with existing data), a document set, and a user instruction (Figure \ref{fig:intro}). The user instruction will provide high-level guidance on which type of action to perform (``Add information about new movies'') and the model is required to extract the relevant information from the documents and update the database accordingly. 
Since each instance deals with a different database, the model must be able to automatically infer which fields (entities, relations, events, attributes) are relevant according to the user instruction and database schema. This is not possible with conventional IE models since the ontology is usually built into the model and each model typically can only handle one type of extraction task.
In addition, the granularity of extraction could be ambiguous when examining the user instruction alone, and analysis of the database content is key to coming up with a precise plan of action. 
Finally, even after the values are extracted from the document, the IE output might need to be normalized before being added to the database. We summarize these key challenges in Figure \ref{fig:challenge}.

To benchmark the \taskname\ task, we introduce an annotated dataset. The dataset includes three high-level categories of instructions: data infilling, row addition, and column addition. It incorporates two sources of databases – simple schemas derived from Wikipedia tables and more complex schemas manually selected from \textsc{Bird}~\cite{Li2023BIRD}. Additionally, the dataset spans various domains, 
to test the generalization ability across different areas. 
The dataset is classified into three difficulty levels – easy, medium, and hard – based on the complexity of the database schema, the length of the source document, and the number of values to update.

While large language models (LLMs) show strong capability of instruction-following and extraction, directly performing the complex task of \taskname\ results in unsatisfactory performance, and scaling up to actual databases would be prohibitively expensive. 
We propose a large language model-based agent framework \ours\ that incorporates multiple IE models as tools.
At its core,  a Planner agent decomposes the user instruction into a code-based plan, which involves data transformations and external calls to specialized models (\textit{e.g.} named entity recognition, relation extraction, entity linking). 
The Analyzer checks the syntax and logic in the program and provides feedback to the planner for iterative improvement.
Simultaneously, the Observer agent is also incorporated to interface with the databases, aiding in database schema analysis, tool selection, and test case generation, thereby ensuring the system's robustness and efficiency.

Our experiments demonstrate the effectiveness of our overall framework and individual components.  We find that equipping the Planner with feedback for self-revision is critical and IE demonstrations from the database help resolve extraction ambiguity and eventually boost extraction quality. 

To conclude, our contributions include (1) we define a new task \taskname\, which is an example of \textit{integration-aware information extraction}, (2) we introduce a new benchmark for \taskname\ with diverse databases and instructions of different difficulty, (3) we design a new LLM agent framework \ours\ (Observe-Plan-Analyze LM). 


\section{The \taskname\ Task}

Our task is defined over a set of (user instruction $I$, database $B$, documents $\mathcal{D}$) instances.
The goal is to automatically update the database ($B \rightarrow B^*$) with new information extracted from a set of text documents to fulfill the user’s request. 
The user instruction $I$ is a natural language sentence indicating the high-level scope and the type of operation. 
The database $B$ contains multiple tables $\mathcal{T}$ that can be filled with pre-existing data entries. The database schema is available to the model, which outlines the data types, constraints, relationships, and integrity rules among different tables. 

Unlike creating a database from scratch,
our task focuses on enriching an existing database $B$
with a collection of text documents $\mathcal{D}$. This setting is more realistic but also more challenging 
since information to be extracted from $\mathcal{D}$ must be aligned with the schema and data in $B$.
\section{\taskname\ Benchmark}

\renewcommand\arraystretch{1.3}
\begin{table}[!h]
\center \footnotesize
\tabcolsep0.065 in
\begin{tabular}{l|cccccc}
\toprule
 & \textbf{\#DB} & \textbf{\#Table} & \textbf{\#Row} & \textbf{\#Col.} & \textbf{\(\Delta \)Value} & \textbf{\#Ins.} \\
\midrule
Wiki & 191 & 1.0 & 116.3 & 5.7 & 6.2 & 195 \\
Bird & 12 & 9.1 & 297K & 55.3 & 17.7 & 45 \\
\bottomrule
\end{tabular}
\caption{Database Comparison. The WikiTable subset of our dataset emphasizes schema diversity whereas the BIRD subset emphasizes database size and complexity. ``Ins.'' denotes the number of instances. More details are provided in the Appendix.}
\label{tab:db_source}
\end{table}

Our benchmark construction starts with selecting a set of databases $\mathcal{B}$ to work with, then finding relevant documents $\mathcal{D}$ and annotating instructions $\mathcal{I}$ and the updated databases $\mathcal{B}'$.

\subsection{Database Selection} 
We use tables from Wikipedia and databases from an existing dataset BIRD. 

We outline the selection criteria and preprocessing procedure for the two sources below: 
\vspace{-0.5ex}
\begin{itemize}
\parskip -0.2ex
    \item \textbf{WikiTables.} The advantage of this data source is that we have a board domain coverage and a natural matching between the tables and documents. 
    We transform these tables into databases by specifying the primary key for each table and performing data cleaning. This involves standardizing column names, removing rows with incomplete values, and excluding descriptive columns that cannot be directly extracted from text. 
    \item \textbf{\textsc{Bird}} \cite{Li2023BIRD}. Databases in BIRD feature multiple tables and complex schemas which introduce dependencies between tables.  We exclude databases that do not contain any column that can be found in public text, focusing instead on those with accessible licenses and real-world applicability. 
\end{itemize}

\subsection{Annotation Process} 
\renewcommand\arraystretch{1.0}
\begin{table}[!h]
\center \footnotesize
\tabcolsep0.05 in
\begin{tabular}{lcccc}
\toprule
\multicolumn{1}{l}{\multirow{2}[1]{*}{\textbf{Data}}} & \multicolumn{3}{c}{\textbf{Task Types}} & \multicolumn{1}{l}{\multirow{2}[1]{*}{\textbf{Total}}}  \\
 & \multicolumn{1}{c}{DI} & \multicolumn{1}{c}{RP} & \multicolumn{1}{c}{CA}  \\
\midrule
\multicolumn{5}{l}{\textbf{User Instruction}} \\
\# Avg. Words & 24.4 & 21.2 & 48.0 & 31.1 \\
\midrule
\multicolumn{5}{l}{\textbf{Source Text}} \\
\# Avg. Docs & 1.0 & 2.0 & 5.2 & 2.7 \\
\# Avg. Words & 1,099.0 & 1,541.9 & 2,739.1 & 1786.5 \\
\midrule
\multicolumn{5}{l}{\textbf{Database}} \\
\# Databases & 73 & 72 & 73 & 203 \\
\# Avg. Tables & 2.6 & 2.6 & 2.4 & 2.5 \\
\# Avg. Rows & 105K & 33K & 29K & 56K \\
\# Avg. Columns & 16.2 & 12.7 & 16.1 & 15.0 \\
\(\Delta \) Values & 1.9 & 11.2 & 12.1 & 8.4 \\
\midrule
\multicolumn{5}{l}{\textbf{Difficulty}} \\
\# Easy & 39 & 10 & 32 & 81 \\
\# Medium & 17  & 38  & 27  & 82 \\
\# Hard  & 25  & 32  & 20 & 77 \\
\midrule
\multicolumn{5}{l}{\textbf{Overall}} \\
\# Data Instances & 81 & 80 & 79 & 240 \\
\# Domains & 23 & 32 & 29 & 45 \\
\bottomrule
\end{tabular}
\caption{Statistics of our dataset. ``DI'', ``RP'', and ``CA'' correspond to three task types, data infilling, row population, and column addition, respectively. ``\#'' indicates the count, ``Avg.'' stands for the average value per instance, and ``\(\Delta \) Values'' represents the number of value changes in the process of database population.}
\label{tab:dataset}
\end{table}

\definecolor{mygreen}{RGB}{231,255,231}
\definecolor{myyellow}{RGB}{255,252,233}
\definecolor{mypurple}{RGB}{230,230,254}

\renewcommand\arraystretch{1.0}
\begin{table*}[!h]
    \centering \footnotesize
    \begin{tabular}{llc }
    \toprule 
       \textbf{Difficulty}  & \multicolumn{1}{c}{\textbf{Criteria}} & \textbf{Number}  \\
    \midrule 
    Easy & \colorbox{mygreen}{\# Table = 1} and \colorbox{myyellow}{\(\Delta\)Values \(\leq\) 10} and \colorbox{mypurple}{Avg. Words \(\leq\) 1k} & 81\\
    Medium & \colorbox{mygreen}{\# Table = 1} and \colorbox{myyellow}{10 < \(\Delta\)Values \(\leq\) 20} and \colorbox{mypurple}{1k < Avg. Words \(\leq\) 2k} & 82 \\
    Hard & \colorbox{mygreen}{\# Table > 1} or \colorbox{myyellow}{\(\Delta\)Values > 20} or \colorbox{mypurple}{Avg. Words > 2k} & 77 \\
    \bottomrule 
   \end{tabular}
   \caption{Criteria of three difficulty levels. ``Avg. Words'' represents the average number of words per document.}
   \label{tab:difficulty}
\end{table*}
We include three general categories of database updates in our benchmark: Data Infilling, Row Population, and Column Addition, 
detailed below.
\begin{itemize}

\item \textbf{Data Infilling} aims to fill in missing values for existing rows. The rows to update are specified by the user or automatically decided by the system. In these rows, the system updates all columns with missing values by default if not specified by users. 

\item \textbf{Row Population} typically adds 1-3 new rows, with the most difficult cases adding up to 10 rows. For each new row, the model should populate as many columns as possible, based on the information available in texts. Otherwise, the default values (as defined in the DB schema) should be inserted. 

\item \textbf{Column Addition} generally adds one to three new columns to a specific table. The instruction should specify the name, meaning, and default value for the new columns, with any special formatting requirements if applicable. The system needs to decide which rows the new values should be linked to. 
\end{itemize}

After selecting a database and the operation category, the human annotator will write a clear and concrete instruction, find related documents, and modify the values in the database to serve as the ground truth for evaluation. 
For detailed guidelines, see Appendix \ref{sec:guidelines}.

\subsection{Statistics}

Our evaluation benchmark includes 240 data instances 
across 203 databases, showcasing a variety of schemas with an average of 2.5 tables (including 56K rows and 15 columns per database on average). The complexity spans from databases with a single table to those with up to 21 tables. Each task in the dataset aims to populate an average of 8.4 values, based on instructions averaging 31 words in length. The overall statistics are shown in Table \ref{tab:dataset}.
The domain distribution within the dataset is well-rounded, featuring significant representations from entertainment (15.4\%), sport (9.2\%), art (8.3\%) and other areas, ensuring comprehensive domain coverage.

The dataset is categorized into three difficulty levels, easy, medium, and hard based on the schema complexity, the size of the required update, and the length of input texts. 
The criteria for determining the difficulty level and difficulty distribution across categories are shown in Table \ref{tab:difficulty}. Note that the size of the required update also positively correlates with the difficulty of the level.


\section{The \ours\ Framework}

 \begin{figure}[!h]
     \centering
     \includegraphics[width=.5\textwidth]{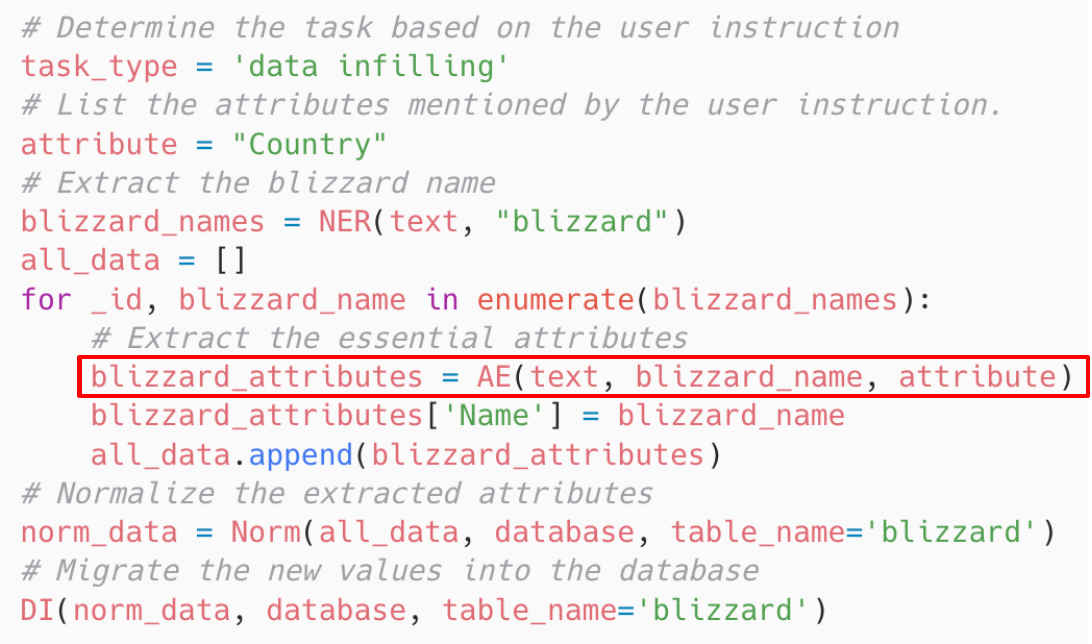}
     \caption{Example of the generated code in one pass, which misconfigures the attribute extraction tool since the tool expects a list of attributes rather than a string.}
     \label{fig:code}
 \end{figure}

We introduce the \ours~(Observe-Plan-Analyze Language Model) framework, 
which starts with observing the target database, then generates the plan of action in code by referring to the database and IE tools and catches errors by both static and dynamic code analysis. 
The \ours\ framework is an instance of an LM agent framework that interacts with a database environment, utilizes IE models as tools, executes actions as code, and improves its plan and tools using feedback from the database.


\definecolor{mypink}{rgb}{.99,.91,.95}

\renewcommand\arraystretch{1.0}
\begin{table*}[t]
    \centering \footnotesize
    \begin{tabular}{lm{34em}}
    \toprule 
    \multicolumn{1}{c}{\textbf{Tools}} & \multicolumn{1}{c}{\textbf{API Signature}} \\
    \midrule
    \rowcolor{mypink!30}
    \multicolumn{2}{c}{\textit{\textbf{Information Extraction}}}\\
    \midrule
    Named Entity Recognition & NER (text: str, type: str) $\rightarrow$ list[str] \\
    Relation Extraction & RE (text: str, head\_e:str, relation: str) $\rightarrow$ list[str] \\
    Attribute Extraction & AE (text: str, entity: str, attribute\_list: list) $\rightarrow$ dict\\
    Text Classification & Classify (text: str, label\_list: list) $\rightarrow$ str \\
    \midrule
    \rowcolor{mypink!30}
    \multicolumn{2}{c}{\textit{\textbf{Database Integration}}}\\
    \midrule
    Entity Linking & Link (data\_entries: list, database: dict, table\_name: str) $\rightarrow$ list\\
    Data Normalization & Norm (data\_entries: list, database: dict, table\_name: str) $\rightarrow$ list\\
    Data Infilling & DI (data\_entry: list, database: dict, table\_name: str) $\rightarrow$ dict \\
    Row Addition & PR (data\_entries: list, database: dict, table\_name: str) $\rightarrow$ dict \\
    Column Addition & AC (data\_entry: list, database: dict, table\_name: str, new\_columns: list) $\rightarrow$ dict \\
       \bottomrule 
    \end{tabular}
    \caption{Tools available to the Planner Agent.}
    \label{tab:tools}
\end{table*}

\subsection{\textsc{Planner}: The Function-Calling Agent}
Since updating a database per user instruction typically requires extracting multiple related fields, instead of asking a model to perform the task end-to-end, 
following the idea of ViperGPT~\cite{Suris2023ViperGPT} and VisProg~\cite{Gupta2022VisualProg}, we first decompose the task into a series of steps represented as code. 
Each of these steps can either be directly executable code or external API calls to a set of IE models (\textit{i.e.} function calls). 

Concretely, the input context $C$ consists of the system prompt $C_0$, the user instruction $I$ and potential input $O$ from the Observer (introduced in Sec. \ref{sec:observer}). The system prompt $C_0$ includes the code APIs for the available tools and in-context examples. 
In the output, inspired by ReAct~\cite{Yao2022ReAct}, we allow the model to generate both code actions and natural language thoughts. The model is free to choose when to output an action and when to output a thought (represented as a comment).


In our work, we define 10 different tools for the Planner agent to use as shown in Table \ref{tab:tools}, spanning standard IE tasks and database primitives. More tool descriptions can be found in Appendix \ref{sec:app_tool}. Figure \ref{fig:code} showcases how the generated code calls for these tools. 
\subsection{\textsc{Analyzer}: The Code Feedback Agent}\label{sec:executor}

In more complicated cases, the Planner often fails to generate the correct actions (code plan) in a single pass (as shown in Figure \ref{fig:code}). 
If the plan execution attempt is unsuccessful, the error message from the code compiler will be provided to the Planner to guide its self-repair process. While this self-repair process has shown to provide some benefit~\cite{Chen2023SelfDebug,leti2023}, it is limited by the quality of the feedback (how well the error message explains the mistake)~\cite{Olausson2023IsSA} and comes at the cost of executing the code multiple times.
In particular, since some of our function calls invoke external models, this process can be very time-consuming. 
To mitigate this problem and improve code quality, we designed the \textsc{Code Analyzer} component, which sits between the Planner and the code compiler, aiming to provide more informative feedback early on.

The input to the Analyzer is the plan of action $A$ (written as code) and the output is the natural language feedback $F$. 
The Analyzer provides feedback of three different types:

\vspace{-0.5ex}
\begin{itemize}
\parskip -0.2ex
    \item \textbf{Syntax error feedback}. Syntax errors might appear in the generated native code (in our case, we use the Python language), or in the API calls. These errors can be directly detected by the interpreter and the Analyzer aims to supplement the error messages with natural language feedback.
    \item \textbf{Runtime and logic error feedback}. Similar to how human programmers debug their code with unit tests, the Analyzer takes a few data samples provided by the Observer and generates test cases by mocking the output from external models. If the output does not match the data samples, the Analyzer will generate another piece of feedback to the Planner. 
    \item \textbf{Database integrity feedback}. Even when the extraction results are correct, we might not be able to update the database successfully due to database constraints. 
    Thus, we implement the functions for the analyzer to checks against duplicating entries and violating database dependency constraints. 
\end{itemize}




 \begin{figure*}
     \centering
     \includegraphics[width=.8\textwidth]{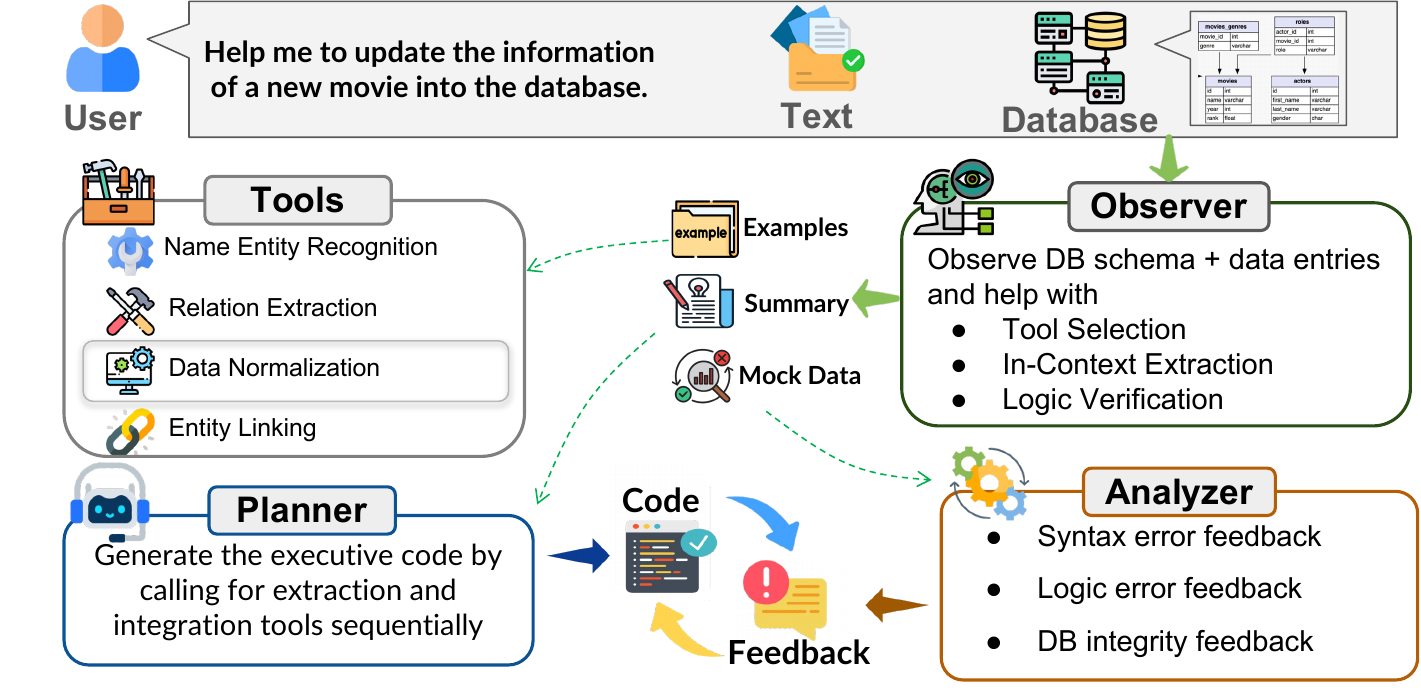}
     \caption{Framework architecture.} 
     \label{fig:model}
 \end{figure*}

\subsection{\textsc{Observer}: The Database Expert Agent}
\label{sec:observer}
In our setting, each database can consist of one or multiple tables, each with its own schema and data entries. Directly providing the whole database as text (by converting to JSON or Markdown code) could dilute the model's attention to other parts of the input prompt. 
We introduce the \textsc{Observer} agent which serves as a bridge between the database environment and other components, including the Planner, Code Analyzer, and the IE models. Speficially: 
\begin{itemize}

\item \textbf{Observer $\rightarrow$ Planner.}
The Observer analyzes the schema and content of the database, identifying crucial aspects behind different columns, such as the format, value range, and semantic meaning. Such insights are summarized into a summary observation $O$, which becomes part of the Planner's input context. 
The observation informs the Planner on selecting the right API call, and whether data normalization is required. 
For instance, for a movie database, the Observer recommends an attribute extractor to extract the movie budget and gross. It might suggest a text classifier to categorize the movie after identifying predefined genres (such as action, comedy, and drama) in the database. 

\item \textbf{Observer $\rightarrow$ IE Models.} 
The type name derived from the database schema alone could be ambiguous for IE models.
For instance, ``location" could refer to a country, city, or a specific area, leading to extractions that may misalign with the database contents. 
To handle this issue, the Observer selectively fetches relevant entries from the database as few-shot demonstrations. 
These demonstrations guide IE tools to understand what to extract and enable in-context learning if applicable. 
For example, if the target column \texttt{Loc} includes values such as \texttt{New York, Los Angeles, Boston}, IE tools can prioritize similar levels of detail (US cities) in the input document. 

\item \textbf{Observer $\rightarrow$ Analyzer.}The Observer selects data for test cases that simulate the user request to help identify logical errors before running the code. 
For instance, if the plan misses the mark on how different tables relate to each other, these simulated tests can highlight those errors early on. 
\end{itemize}
\section{Experiments}

\renewcommand\arraystretch{1}
\begin{table*}[t]
\center \footnotesize
\tabcolsep0.11 in
\begin{tabular}{lccccccccc}
\toprule
\multicolumn{1}{l}{\multirow{2}[1]{*}{\textbf{Models}}} & \multicolumn{3}{c}{\textbf{Difficulty}}
 & \multicolumn{3}{c}{\textbf{Task Type}} & \multicolumn{2}{c}{\textbf{DB Source}} & \multicolumn{1}{l}{\multirow{2}[1]{*}{\textbf{Overall}}} \\
 & \multicolumn{1}{c}{Easy} & \multicolumn{1}{c}{Medium} & \multicolumn{1}{c}{Hard} & \multicolumn{1}{c}{DI} & \multicolumn{1}{c}{RP} & \multicolumn{1}{c}{CA} & \multicolumn{1}{c}{Wiki} & \multicolumn{1}{c}{Bird}  \\
 
\cmidrule(lr){1-1} \cmidrule(lr){2-4} \cmidrule(lr){5-7} \cmidrule(lr){8-9} \cmidrule(lr){10-10}

\multicolumn{9}{l}{\textbf{Planner}} \\
\textit{Template} & 11.19 & 8.21 & 0.35 & 7.23 & 12.88 & 0.00 & 8.07 & 0.0 & \cellcolor[rgb]{ .851,  .851,  .851} 6.73 \\

\textit{One-shot} & 18.77 & 14.45 & 2.51 & 16.25 & 12.83 & 7.04 & 14.85  & 0.04 & \cellcolor[rgb]{ .851,  .851,  .851} 12.08  \\


\midrule
\multicolumn{9}{l}{\textbf{IE tools}} \\
\textit{GPT $\rightarrow$ Small Models} & 20.63 & 21.24 & 13.44 & 24.21 & 9.84 & 21.42 & 19.10 & 15.27 & \cellcolor[rgb]{ .851,  .851,  .851} 18.50  \\

\midrule
\multicolumn{9}{l}{\textbf{Observer}} \\

\textit{\quad - Observer } & 23.24 & 22.87 & 18.49 & 27.05 & 14.72 & 22.95 & 23.17 & 14.74 & \cellcolor[rgb]{ .851,  .851,  .851} 21.59  \\

\textit{\quad - DB Analysis } & 38.55 & 26.78 & 22.70 & 30.55 & 32.48 & 24.76 & 31.83 & 18.25 & \cellcolor[rgb]{ .851,  .851,  .851} 29.29  \\
\textit{\quad - IE Demonstration} & 25.96 & 28.85 & 15.38 & 26.18 & 20.00 & 24.30 & 24.15 & 20.24 & \cellcolor[rgb]{ .851,  .851,  .851} 23.50  \\
\textit{\quad - Simulated Test } & 42.42 & 33.79 & 22.00 & 38.93 & 32.77 & 26.30 & 36.90 & 11.81 & \cellcolor[rgb]{ .851,  .851,  .851} 32.72 \\

\midrule
\multicolumn{9}{l}{\textbf{Full Model}} \\ \textsc{\ours} & \textbf{42.44} & \textbf{36.91} & \textbf{23.21} & \textbf{38.85} & \textbf{37.01} & \textbf{26.34} &  \textbf{36.97} & \textbf{21.74} & \cellcolor[rgb]{ .851,  .851,  .851} \textbf{34.11}  \\

\bottomrule
\end{tabular}
\caption{Experiment results. The metric is $\rm{F}_1$ (\%) of the exact matching score.}
\label{tab:main_exp}
\end{table*}

\subsection{Experiment Setting}
\paragraph{Evaluation Metric}
We evaluate the models by comparing the database before and after updating, checking the difference $\Delta {B}$.
We represent the database entries in $\Delta {B}$ as a set of structured tuples $T$, following the form of \textit{(table name, primary key, primary key value, column name, value of that column)}.
Each updated entry is ruled as correct if all the fields match the ground truth (Exact Match). Then we compute F1 over all entries in the database update $\Delta {B}$.  The reported metric is \textbf{macro-F1}, in other words, F1 averaged over each (instruction, document set, database) instance. 

\paragraph{Implementation}
The planner and observer agents in our \ours\ framework are powered by the GPT4 language model \texttt{gpt-4-1106-preview}. 
The maximum number for the Planner to revise plans is 10. The whole process can restart for 2 times at most after failure.
For the tool library, we emulate the models with GPT for named entity recognition, relation extraction, attribute extraction, text classification and data normalization. In addition, we adopt an existing entity linking model GENRE \cite{de-cao-etal-2022-multilingual}. 
More details and prompts can be found in Appendix \ref{sec:app_implementation}. 


\subsection{Experiment Results}
We show our evaluation results in Table \ref{tab:main_exp}. 
We have the following observations:

\textbf{(1) \ours\ vs Template: Dynamic plans are necessary.} Our first baseline replaces the planner with a static template which first uses an entity extraction tool to extract the primary key of the table (selected by the Observer) and then goes on to extract each column by using the attribute extraction tool. 
In our \taskname\ setting, the diversity of schemas and instruction led to a sharp decline in performance over all slices of the dataset.

\textbf{(2) \ours\ vs One-Shot: Feedback improves plan quality.}
The plan generated by the Planner in its first attempt is often error-prone. By utilizing the Analyzer and allowing the Planner to make multiple rounds of revision, we can achieve a large gain in performance.

\textbf{(3) Smaller IE models are less capable of zero-shot/few-shot learning.} We defined our IE tools as few-shot learners without a fixed ontology since the type names are provided as part of the input. In this setting, we find that LLM-emulated IE models work better than fine-tuned smaller-scaled models \cite{li2022ultra, sainz2021label, lyu2021zero, gera2022zero}, due to the strong in-context learning ability of LLMs. However, since many of the errors are from the extraction stage (see Section \ref{sec:error}), perhaps a better choice would be to automatically route the API call to an LLM or specialized model based on the requested type. 
    
\textbf{(4) The Observer is most helpful by providing IE demonstrations.} Among the multiple functions of the Observer component, we see that selecting a few values from the target table (or target column) to serve as few-shot demonstrations significantly contribute to the final performance. This partially resolves the challenge of \textit{extraction ambiguity}. 

\textbf{(5) The Observer is more useful when the database is large and complex.} When the number of tables is larger and there are more dependencies between tables, using the Observer to generate simulated test cases helps improve the plan quality.

 \begin{figure}[!h]
     \centering
     \includegraphics[width=0.45\textwidth]{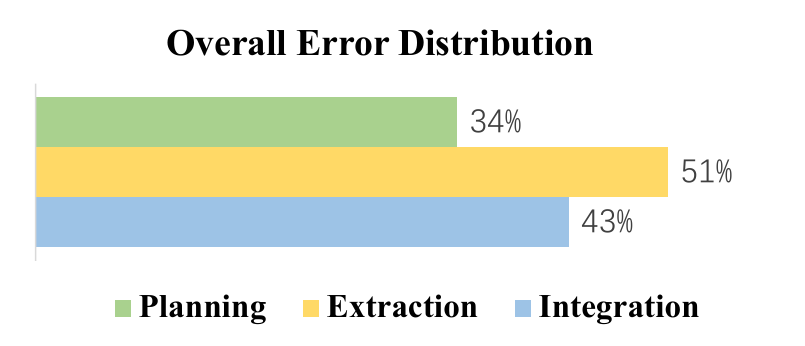}
     \caption{Error distribution of the whole framework.}
     \label{fig:error1}
 \end{figure}
\subsection{Case study}

 \begin{figure*}[!h]
     \centering
     \includegraphics[width=\textwidth]{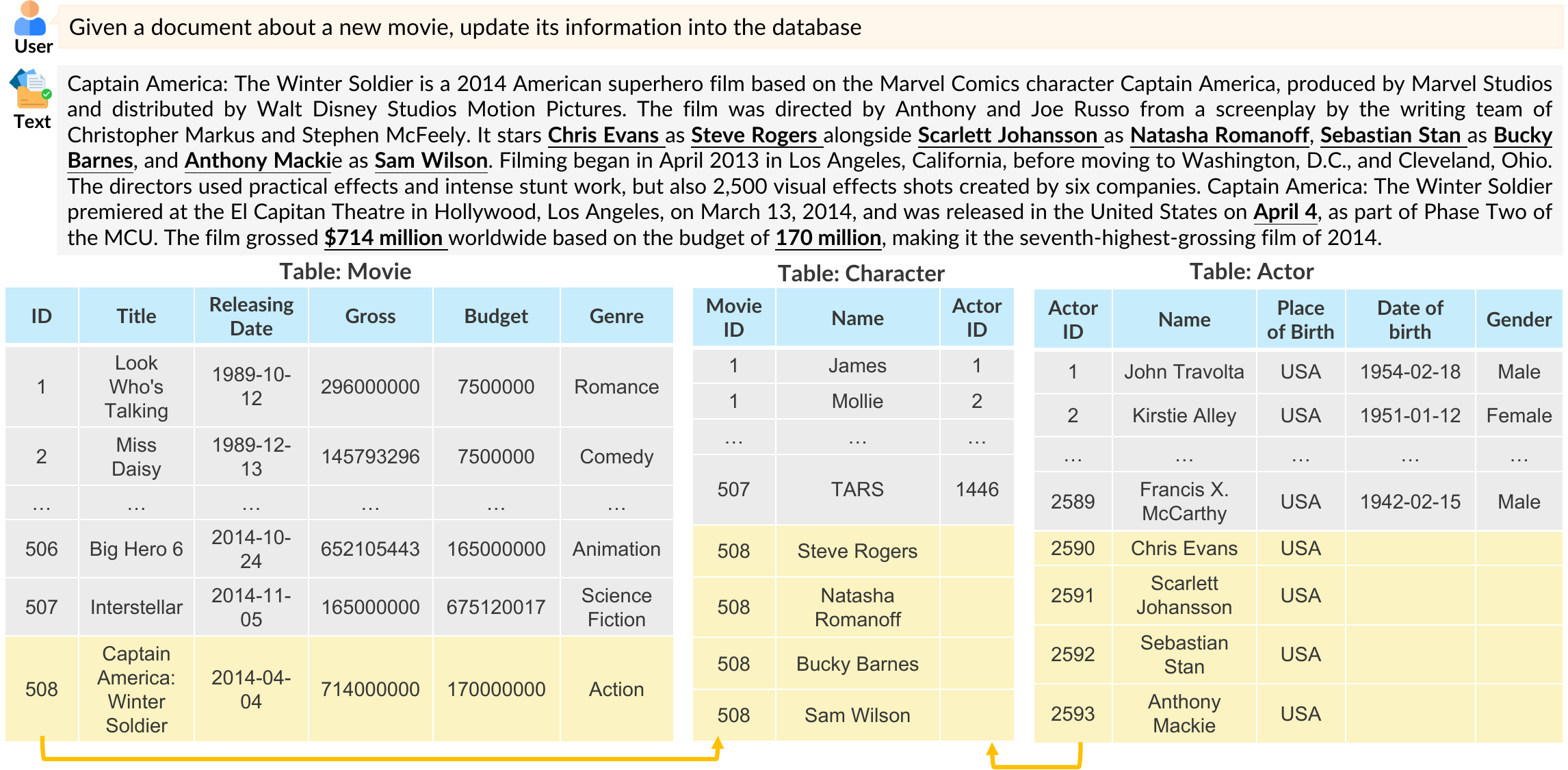}
     \caption{Case study of a row population task on a movie database with three tables. The model successfully extracts relevant information from the document and normalizes the data to conform with the database (\texttt{Movie.Releasing Date, Movie.Gross}). However, the model still struggles to deal with table dependencies (\texttt{Character.ActorID} should refer to \texttt{Actor.ActorID}) and occasionally hallucinates values (\texttt{Actor.PlaceOfBirth}). }
     \label{fig:case}
 \end{figure*}

Figure \ref{fig:case} shows a case of updating a movie database. 
The database includes three tables (\texttt{Movie, Actor, Character}) with two foreign key dependencies. To fulfill the user's instruction, the model must add new entries in all three tables while ensuring database integrity. 
The planner effectively identified extraction targets across tables and generated appropriate function calls, for instance, using an attribute extractor for the \texttt{Gross} column and employing a classifier for the \texttt{Genre} column.
However, it mistakenly attempted to find  \texttt{Character.ActorID} by the character name, suggesting a need for better database understanding.
Extraction tools accurately identify the desired data from text, though they erroneously make up the actors' ``birthplace'', pointing to the need to improve extraction faithfulness, possibly through verification mechanisms.
Integration is successful, effectively normalizing data formats like release dates and movie gross, and correctly assigning primary keys to new entries.

\subsection{Error Analysis}
\label{sec:error}

We analyzed 100 wrong cases by classifying them based on the stage of error: planning, extraction, or integration. Our findings in Figure \ref{fig:error1} show that 34\% of errors occurred during the planning phase, particularly in databases with complex schemas and multiple tables, aligning with our findings in Table \ref{tab:main_exp}. Errors in planning were notably more frequent in row addition tasks due to their likelihood of involving multiple tables. 
Extraction errors are most common, driven by dependencies in the extraction process, such as the need for named entity recognition before relation extraction, leading to error accumulation. 
In the integration phase, the majority of errors are related to data infilling and column addition, with the bottleneck being entity linking. A more detailed analysis of each stage can be found in Appendix \ref{sec:app_error}. 

\section{Related Work}
\paragraph{LLMs for Databases}
The majority of work involving the application of LLMs to databases focuses on the text-to-SQL task~\cite{yu-etal-2018-spider,Li2023BIRD,Liu2023AgentBench} which does not involve extraction. 
Recently, LLMs with few-shot examples have shown to  outperform fine-tuned smaller models \cite{Pourreza2023DINSQL,zhang-etal-2023-act,Sun2023SQLPaLM}.

The structured view generation task in ~\cite{Arora2023LanguageME} is closest to our \taskname\ setting. However, their task only requires extraction from semi-structured documents, without the need for integration with existing tables. We note that when the target database is of the form of (entity name, entity attribute), our task resembles knowledge base population (KBP)~\cite{Getman2018LayingTG}. The key difference lies in the fact that our task requires dynamic adaptation to diverse databases.

\paragraph{Tool Learning in LLMs}
Tool learning, or function-calling, has emerged as a promising approach to extend the capability of large language models. In the tool learning paradigm, certain tools (such as internet search, a calculator, and image generation models) are provided to the LLM.
 Tool learning can be enabled by prompting with function definitions or specialized fine-tuning~\cite{Schick2023Toolformer, Tang2023ToolAlpaca, Patil2023Gorilla,Zeng2023AgentTuning}\footnote{It is speculated that GPT-4 has been fine-tuned to support function calling.}. We refer the reader to \cite{Qin2023ToolSurvey} for a comprehensive survey.
 In the \ours\ framework, IE models act as external tools to the Planner. 
Unlike prior work that leaves all the work to the Planner and assumes that tools as provided as-is, our Observer provides essential insight into the database to support tool selection and selects demonstrations to assist the IE tools.

\section{Conclusion and Future Work}
In this paper, we propose the new task \taskname\, which updates a given database using values extracted from a document set following user instructions. \taskname\ presents unique challenges in dynamic adaptation
, extraction ambiguity 
and integrity requirements.
We present a new benchmark and a new LLM agent framework \ours\ for this task. 
In particular, \ours\ features 3 components: the Planner, Analyzer, and Observer.
We show that \ours\ substantially improves update effectiveness over directly generating the plan, with significant gains coming from using demonstrations from the Observer for IE models. 

At this point, we have only considered inserting values that do not previously exist into the database. However, conflict resolution has been a long-standing issue in the integration of databases, and we also foresee similar challenges in the integration of documents and databases. 
In this case, we need to consider the reliability of the document and the confidence of the extraction model.

\section*{Limitations}
The success of our agent framework \ours\ relies heavily on the instruction-following and tool-using ability of the language model. As a proof-of-concept for our new task, we used the most capable LLM GPT4 at the time of writing. Benchmarking different base LLMs would provide extra insight to how the framework generalizes. In addition, We have only experimented with a small set of IE models as tools (one for each API call) and could have been extended to a larger repository of open-source models such as that in \citet{Shen2023HuggingGPT}.

\section*{Acknowledgements}
Thanks to Yu Su at the Ohio State University and Tanay Dixit at Universtiy of Illinois at Urbana-Champaign for their enlightening discussions. 
Research was supported in part by US DARPA KAIROS Program No. FA8750-19-2-1004 and INCAS Program No. HR001121C0165, National Science Foundation IIS-19-56151, and the Molecule Maker Lab Institute: An AI Research Institutes program supported by NSF under Award No. 2019897, and the Institute for Geospatial Understanding through an Integrative Discovery Environment (I-GUIDE) by NSF under Award No. 2118329. 

\bibliography{custom}

\begin{thebibliography}{30}
\expandafter\ifx\csname natexlab\endcsname\relax\def\natexlab#1{#1}\fi

\bibitem[{Arora et~al.(2023)Arora, Yang, Eyuboglu, Narayan, Hojel, Trummer, and R{\'e}}]{Arora2023LanguageME}
Simran Arora, Brandon Yang, Sabri Eyuboglu, Avanika Narayan, Andrew Hojel, Immanuel Trummer, and Christopher R{\'e}. 2023.
\newblock \href {https://api.semanticscholar.org/CorpusID:258212828} {Language models enable simple systems for generating structured views of heterogeneous data lakes}.
\newblock \emph{ArXiv}, abs/2304.09433.

\bibitem[{Chen et~al.(2023)Chen, Lin, Sch{\"a}rli, and Zhou}]{Chen2023SelfDebug}
Xinyun Chen, Maxwell Lin, Nathanael Sch{\"a}rli, and Denny Zhou. 2023.
\newblock \href {https://api.semanticscholar.org/CorpusID:258059885} {Teaching large language models to self-debug}.
\newblock \emph{ArXiv}, abs/2304.05128.

\bibitem[{De~Cao et~al.(2022)De~Cao, Wu, Popat, Artetxe, Goyal, Plekhanov, Zettlemoyer, Cancedda, Riedel, and Petroni}]{de-cao-etal-2022-multilingual}
Nicola De~Cao, Ledell Wu, Kashyap Popat, Mikel Artetxe, Naman Goyal, Mikhail Plekhanov, Luke Zettlemoyer, Nicola Cancedda, Sebastian Riedel, and Fabio Petroni. 2022.
\newblock \href {https://doi.org/10.1162/tacl_a_00460} {Multilingual autoregressive entity linking}.
\newblock \emph{Transactions of the Association for Computational Linguistics}, 10:274--290.

\bibitem[{Ding et~al.(2021)Ding, Xu, Chen, Wang, Han, Xie, Zheng, and Liu}]{ding2021few}
Ning Ding, Guangwei Xu, Yulin Chen, Xiaobin Wang, Xu~Han, Pengjun Xie, Haitao Zheng, and Zhiyuan Liu. 2021.
\newblock Few-nerd: A few-shot named entity recognition dataset.
\newblock In \emph{Proceedings of the 59th Annual Meeting of the Association for Computational Linguistics and the 11th International Joint Conference on Natural Language Processing (Volume 1: Long Papers)}, pages 3198--3213.

\bibitem[{Gera et~al.(2022)Gera, Halfon, Shnarch, Perlitz, Dor, and Slonim}]{gera2022zero}
Ariel Gera, Alon Halfon, Eyal Shnarch, Yotam Perlitz, Liat~Ein Dor, and Noam Slonim. 2022.
\newblock Zero-shot text classification with self-training.
\newblock In \emph{Proceedings of the 2022 Conference on Empirical Methods in Natural Language Processing}, pages 1107--1119.

\bibitem[{Getman et~al.(2018)Getman, Ellis, Strassel, Song, and Tracey}]{Getman2018LayingTG}
Jeremy Getman, Joe Ellis, Stephanie Strassel, Zhiyi Song, and Jennifer Tracey. 2018.
\newblock \href {https://api.semanticscholar.org/CorpusID:21713497} {Laying the groundwork for knowledge base population: Nine years of linguistic resources for tac kbp}.
\newblock In \emph{International Conference on Language Resources and Evaluation}.

\bibitem[{Gupta and Kembhavi(2023)}]{Gupta2022VisualProg}
Tanmay Gupta and Aniruddha Kembhavi. 2023.
\newblock \href {https://api.semanticscholar.org/CorpusID:253734854} {Visual programming: Compositional visual reasoning without training}.
\newblock \emph{2023 IEEE/CVF Conference on Computer Vision and Pattern Recognition (CVPR)}, pages 14953--14962.

\bibitem[{Li et~al.(2022)Li, Yin, and Chen}]{li2022ultra}
Bangzheng Li, Wenpeng Yin, and Muhao Chen. 2022.
\newblock Ultra-fine entity typing with indirect supervision from natural language inference.
\newblock \emph{Transactions of the Association for Computational Linguistics}, 10:607--622.

\bibitem[{Li et~al.(2023)Li, Hui, Qu, Li, Yang, Li, Wang, Qin, Cao, Geng, Huo, Ma, Chang, Huang, Cheng, and Li}]{Li2023BIRD}
Jinyang Li, Binyuan Hui, Ge~Qu, Binhua Li, Jiaxi Yang, Bowen Li, Bailin Wang, Bowen Qin, Rongyu Cao, Ruiying Geng, Nan Huo, Chenhao Ma, Kevin~C. Chang, Fei Huang, Reynold Cheng, and Yongbin Li. 2023.
\newblock \href {https://api.semanticscholar.org/CorpusID:258547040} {Can llm already serve as a database interface? a big bench for large-scale database grounded text-to-sqls}.
\newblock \emph{ArXiv}, abs/2305.03111.

\bibitem[{Li et~al.(2021)Li, Ji, and Han}]{Li2021doclevel}
Sha Li, Heng Ji, and Jiawei Han. 2021.
\newblock Document-level event argument extraction by conditional generation.
\newblock In \emph{Proc. The 2021 Conference of the North American Chapter of the Association for Computational Linguistics - Human Language Technologies (NAACL-HLT2021)}.

\bibitem[{Liu et~al.(2023)Liu, Yu, Zhang, Xu, Lei, Lai, Gu, Gu, Ding, Men, Yang, Zhang, Deng, Zeng, Du, Zhang, Shen, Zhang, Su, Sun, Huang, Dong, and Tang}]{Liu2023AgentBench}
Xiao Liu, Hao Yu, Hanchen Zhang, Yifan Xu, Xuanyu Lei, Hanyu Lai, Yu~Gu, Yuxian Gu, Hangliang Ding, Kai Men, Kejuan Yang, Shudan Zhang, Xiang Deng, Aohan Zeng, Zhengxiao Du, Chenhui Zhang, Shengqi Shen, Tianjun Zhang, Yu~Su, Huan Sun, Minlie Huang, Yuxiao Dong, and Jie Tang. 2023.
\newblock \href {https://api.semanticscholar.org/CorpusID:260682249} {Agentbench: Evaluating llms as agents}.
\newblock \emph{ArXiv}, abs/2308.03688.

\bibitem[{Lyu et~al.(2021)Lyu, Zhang, Sulem, and Roth}]{lyu2021zero}
Qing Lyu, Hongming Zhang, Elior Sulem, and Dan Roth. 2021.
\newblock Zero-shot event extraction via transfer learning: Challenges and insights.
\newblock In \emph{Proceedings of the 59th Annual Meeting of the Association for Computational Linguistics and the 11th International Joint Conference on Natural Language Processing (Volume 2: Short Papers)}, pages 322--332.

\bibitem[{Muhammad et~al.(2020)Muhammad, Kearney, Gamble, Coenen, and Williamson}]{DBLP:conf/dexaw/MuhammadKGCW20}
Iqra Muhammad, Anna Kearney, Carrol Gamble, Frans Coenen, and Paula Williamson. 2020.
\newblock \href {https://doi.org/10.1007/978-3-030-59028-4\_10} {Open information extraction for knowledge graph construction}.
\newblock In \emph{Database and Expert Systems Applications - {DEXA} 2020 International Workshops BIOKDD, {IWCFS} and MLKgraphs, Bratislava, Slovakia, September 14-17, 2020, Proceedings}, volume 1285 of \emph{Communications in Computer and Information Science}, pages 103--113. Springer.

\bibitem[{Olausson et~al.(2024)Olausson, Inala, Wang, Gao, and Solar-Lezama}]{Olausson2023IsSA}
Theo~X. Olausson, Jeevana~Priya Inala, Chenglong Wang, Jianfeng Gao, and Armando Solar-Lezama. 2024.
\newblock Is self-repair a silver bullet for code generation?

\bibitem[{Patil et~al.(2023)Patil, Zhang, Wang, and Gonzalez}]{Patil2023Gorilla}
Shishir~G. Patil, Tianjun Zhang, Xin Wang, and Joseph~E. Gonzalez. 2023.
\newblock \href {https://api.semanticscholar.org/CorpusID:258865184} {Gorilla: Large language model connected with massive apis}.
\newblock \emph{ArXiv}, abs/2305.15334.

\bibitem[{Pourreza and Rafiei(2023)}]{Pourreza2023DINSQL}
Mohammad~Reza Pourreza and Davood Rafiei. 2023.
\newblock \href {https://api.semanticscholar.org/CorpusID:258291425} {Din-sql: Decomposed in-context learning of text-to-sql with self-correction}.
\newblock \emph{Neurips}.

\bibitem[{Qin et~al.(2023)Qin, Hu, Lin, Chen, Ding, Cui, Zeng, Huang, Xiao, Han, Fung, Su, Wang, Qian, Tian, Zhu, Liang, Shen, Xu, Zhang, Ye, Li, Tang, Yi, Zhu, Dai, Yan, Cong, Lu, Zhao, Huang, Yan, Han, Sun, Li, Phang, Yang, Wu, Ji, Liu, and Sun}]{Qin2023ToolSurvey}
Yujia Qin, Shengding Hu, Yankai Lin, Weize Chen, Ning Ding, Ganqu Cui, Zheni Zeng, Yufei Huang, Chaojun Xiao, Chi Han, Yi~Ren Fung, Yusheng Su, Huadong Wang, Cheng Qian, Runchu Tian, Kunlun Zhu, Shi Liang, Xingyu Shen, Bokai Xu, Zhen Zhang, Yining Ye, Bo~Li, Ziwei Tang, Jing Yi, Yu~Zhu, Zhenning Dai, Lan Yan, Xin Cong, Ya-Ting Lu, Weilin Zhao, Yuxiang Huang, Jun-Han Yan, Xu~Han, Xian Sun, Dahai Li, Jason Phang, Cheng Yang, Tongshuang Wu, Heng Ji, Zhiyuan Liu, and Maosong Sun. 2023.
\newblock \href {https://api.semanticscholar.org/CorpusID:258179336} {Tool learning with foundation models}.
\newblock \emph{ArXiv}, abs/2304.08354.

\bibitem[{Sainz et~al.(2021)Sainz, de~Lacalle, Labaka, Barrena, and Agirre}]{sainz2021label}
Oscar Sainz, Oier~Lopez de~Lacalle, Gorka Labaka, Ander Barrena, and Eneko Agirre. 2021.
\newblock Label verbalization and entailment for effective zero and few-shot relation extraction.
\newblock In \emph{Proceedings of the 2021 Conference on Empirical Methods in Natural Language Processing}, pages 1199--1212.

\bibitem[{Schick et~al.(2023)Schick, Dwivedi-Yu, Dess{\`i}, Raileanu, Lomeli, Zettlemoyer, Cancedda, and Scialom}]{Schick2023Toolformer}
Timo Schick, Jane Dwivedi-Yu, Roberto Dess{\`i}, Roberta Raileanu, Maria Lomeli, Luke Zettlemoyer, Nicola Cancedda, and Thomas Scialom. 2023.
\newblock \href {https://api.semanticscholar.org/CorpusID:256697342} {Toolformer: Language models can teach themselves to use tools}.
\newblock \emph{Neurips}.

\bibitem[{Shen et~al.(2023)Shen, Song, Tan, Li, Lu, and Zhuang}]{Shen2023HuggingGPT}
Yongliang Shen, Kaitao Song, Xu~Tan, Dong~Sheng Li, Weiming Lu, and Yue~Ting Zhuang. 2023.
\newblock \href {https://api.semanticscholar.org/CorpusID:257833781} {Hugginggpt: Solving ai tasks with chatgpt and its friends in hugging face}.
\newblock \emph{ArXiv}, abs/2303.17580.

\bibitem[{Sun et~al.(2023)Sun, Arik, Nakhost, Dai, Sinha, Yin, and Pfister}]{Sun2023SQLPaLM}
Ruoxi Sun, Sercan~{\"O}. Arik, Hootan Nakhost, Hanjun Dai, Rajarishi Sinha, Pengcheng Yin, and Tomas Pfister. 2023.
\newblock \href {https://api.semanticscholar.org/CorpusID:258999853} {Sql-palm: Improved large language model adaptation for text-to-sql}.
\newblock \emph{ArXiv}, abs/2306.00739.

\bibitem[{Sur'is et~al.(2023)Sur'is, Menon, and Vondrick}]{Suris2023ViperGPT}
D'idac Sur'is, Sachit Menon, and Carl Vondrick. 2023.
\newblock \href {https://api.semanticscholar.org/CorpusID:257505358} {Vipergpt: Visual inference via python execution for reasoning}.
\newblock \emph{2023 IEEE/CVF International Conference on Computer Vision (ICCV)}, pages 11854--11864.

\bibitem[{Tang et~al.(2023)Tang, Deng, Lin, Han, Liang, Cao, and Sun}]{Tang2023ToolAlpaca}
Qiaoyu Tang, Ziliang Deng, Hongyu Lin, Xianpei Han, Qiao Liang, Boxi Cao, and Le~Sun. 2023.
\newblock \href {https://api.semanticscholar.org/CorpusID:259108190} {Toolalpaca: Generalized tool learning for language models with 3000 simulated cases}.
\newblock \emph{ArXiv}, abs/2306.05301.

\bibitem[{Wan et~al.(2023)Wan, Cheng, Mao, Liu, Song, Li, and Kurohashi}]{wan2023gpt}
Zhen Wan, Fei Cheng, Zhuoyuan Mao, Qianying Liu, Haiyue Song, Jiwei Li, and Sadao Kurohashi. 2023.
\newblock Gpt-re: In-context learning for relation extraction using large language models.
\newblock In \emph{Proceedings of the 2023 Conference on Empirical Methods in Natural Language Processing}, pages 3534--3547.

\bibitem[{Wang et~al.(2023)Wang, Peng, Jabbarvand, and Ji}]{leti2023}
Xingyao Wang, Hao Peng, Reyhaneh Jabbarvand, and Heng Ji. 2023.
\newblock Leti: Learning to generate from textual interactions.
\newblock In \emph{arxiv}.

\bibitem[{Weischedel et~al.(2013)Weischedel, Palmer, Marcus, Hovy, Pradhan, Ramshaw, Xue, Taylor, Kaufman, Franchini et~al.}]{weischedel2013ontonotes}
Ralph Weischedel, Martha Palmer, Mitchell Marcus, Eduard Hovy, Sameer Pradhan, Lance Ramshaw, Nianwen Xue, Ann Taylor, Jeff Kaufman, Michelle Franchini, et~al. 2013.
\newblock \href {https://catalog.ldc.upenn.edu/LDC2013T19} {Ontonotes release 5.0 ldc2013t19}.
\newblock \emph{Linguistic Data Consortium, Philadelphia, PA}, 23.

\bibitem[{Yao et~al.(2022)Yao, Zhao, Yu, Du, Shafran, Narasimhan, and Cao}]{Yao2022ReAct}
Shunyu Yao, Jeffrey Zhao, Dian Yu, Nan Du, Izhak Shafran, Karthik Narasimhan, and Yuan Cao. 2022.
\newblock \href {https://api.semanticscholar.org/CorpusID:252762395} {React: Synergizing reasoning and acting in language models}.
\newblock \emph{ArXiv}, abs/2210.03629.

\bibitem[{Yu et~al.(2018)Yu, Zhang, Yang, Yasunaga, Wang, Li, Ma, Li, Yao, Roman, Zhang, and Radev}]{yu-etal-2018-spider}
Tao Yu, Rui Zhang, Kai Yang, Michihiro Yasunaga, Dongxu Wang, Zifan Li, James Ma, Irene Li, Qingning Yao, Shanelle Roman, Zilin Zhang, and Dragomir Radev. 2018.
\newblock \href {https://doi.org/10.18653/v1/D18-1425} {{S}pider: A large-scale human-labeled dataset for complex and cross-domain semantic parsing and text-to-{SQL} task}.
\newblock In \emph{Proceedings of the 2018 Conference on Empirical Methods in Natural Language Processing}, pages 3911--3921, Brussels, Belgium. Association for Computational Linguistics.

\bibitem[{Zeng et~al.(2023)Zeng, Liu, Lu, Wang, Liu, Dong, and Tang}]{Zeng2023AgentTuning}
Aohan Zeng, Mingdao Liu, Rui Lu, Bowen Wang, Xiao Liu, Yuxiao Dong, and Jie Tang. 2023.
\newblock \href {https://api.semanticscholar.org/CorpusID:264306101} {Agenttuning: Enabling generalized agent abilities for llms}.
\newblock \emph{ArXiv}, abs/2310.12823.

\bibitem[{Zhang et~al.(2023)Zhang, Cao, Chen, Xu, and Yu}]{zhang-etal-2023-act}
Hanchong Zhang, Ruisheng Cao, Lu~Chen, Hongshen Xu, and Kai Yu. 2023.
\newblock \href {https://doi.org/10.18653/v1/2023.findings-emnlp.227} {{ACT}-{SQL}: In-context learning for text-to-{SQL} with automatically-generated chain-of-thought}.
\newblock In \emph{Findings of the Association for Computational Linguistics: EMNLP 2023}, pages 3501--3532, Singapore. Association for Computational Linguistics.

\end{thebibliography}

\appendix

\section{Benchmark Annotation Guidelines}

\label{sec:guidelines}
\begin{figure*}
     \centering
     \includegraphics[width=\textwidth]{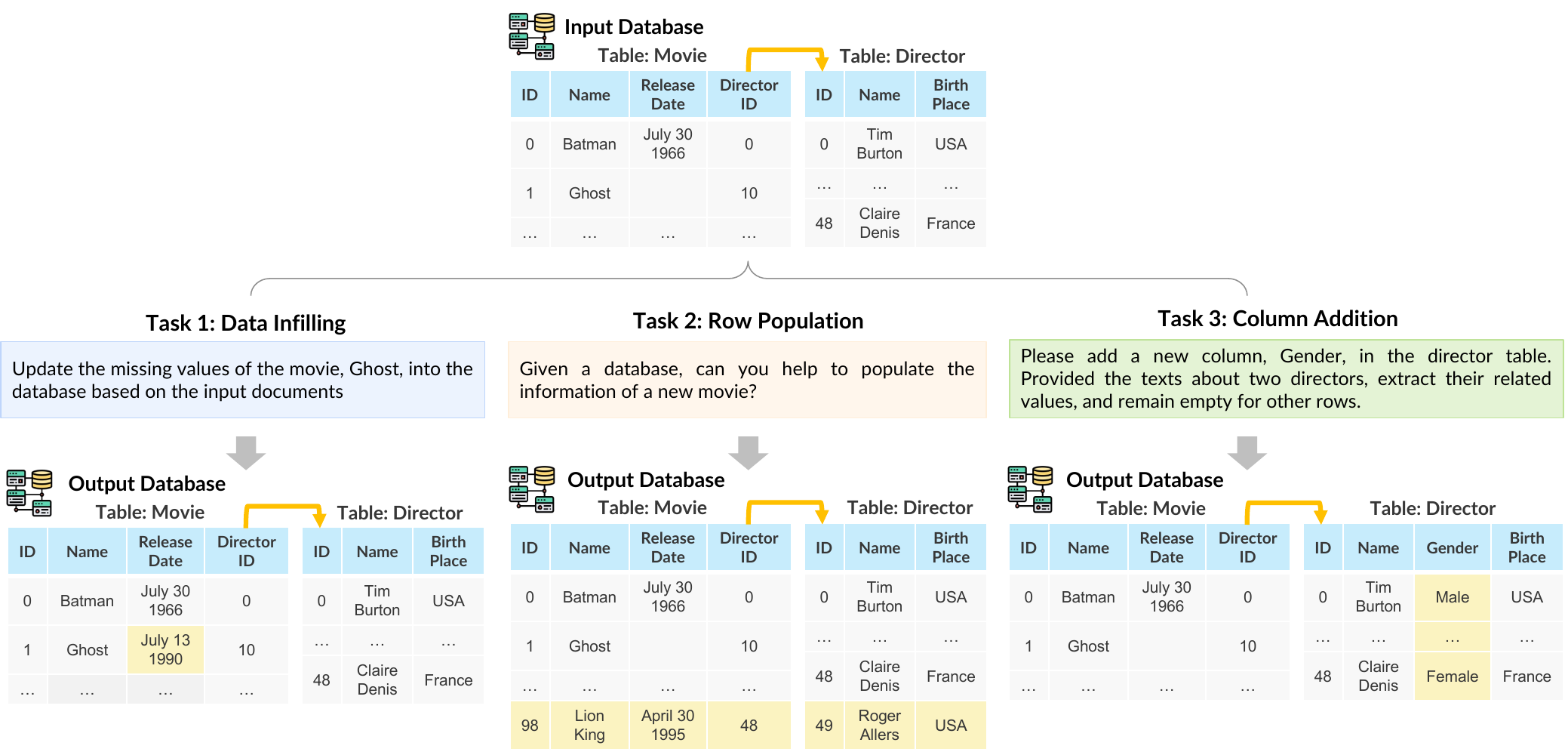}
     \caption{Examples of three task types including data infilling, row population, and column addition. }
     \label{fig:task}
 \end{figure*}
 \renewcommand\arraystretch{1.0}
\begin{table}[t]
\center \footnotesize
\tabcolsep0.1 in
\begin{tabular}{lccc}
\toprule
\multicolumn{1}{l}{\multirow{2}[1]{*}{\textbf{Statistics}}} & \multicolumn{2}{c}{\textbf{Task Types}} & \multicolumn{1}{l}{\multirow{2}[1]{*}{\textbf{Total}}}  \\
 & \multicolumn{1}{c}{Wiki} & \multicolumn{1}{c}{Bird}   \\
\midrule
\multicolumn{4}{l}{\textbf{User Instruction}} \\
\# Avg. Words & 29.7 & 37.5 & 31.1 \\
\midrule
\multicolumn{4}{l}{\textbf{Source Text}}\\
\# Avg. Documents & 2.8 & 2.4 & 2.7 \\
\# Avg. Words & 2,123.8 & 325.0 & 1786.5 \\
\midrule
\multicolumn{4}{l}{\textbf{Database}}\\
\# Databases & 191 & 12 & 203 \\
\# Avg. Tables & 1.0 & 9.1 & 2.5 \\
\# Avg. Rows & 116.3 & 297K & 56K \\
\# Avg. Columns & 5.7 & 55.3 & 15.0 \\
\(\Delta \)Values & 6.2 & 17.7 & 8.4 \\
\midrule
\multicolumn{4}{l}{\textbf{Overall}}\\
\# Domains & 42 & 9 & 45 \\
\# Easy & 81 & 0 & 81 \\
\# Medium & 82 & 0 & 82 \\
\# Hard & 32 & 45 & 77 \\
\# Instances & 195 & 45 & 240 \\
\bottomrule
\end{tabular}
\caption{Statistics of our dataset divided by database source. ``Wiki'' and ``Bird'' correspond to different database sources. ``\#'' indicates the count, ``Avg.'' stands for the average value per instance, and ``\(\Delta \)Values'' represents the number of value changes in the database following the completion of integration.}
\label{tab:dataset_dbsource}
\end{table}

  \begin{figure}
     \centering
     \includegraphics[width=0.5\textwidth]{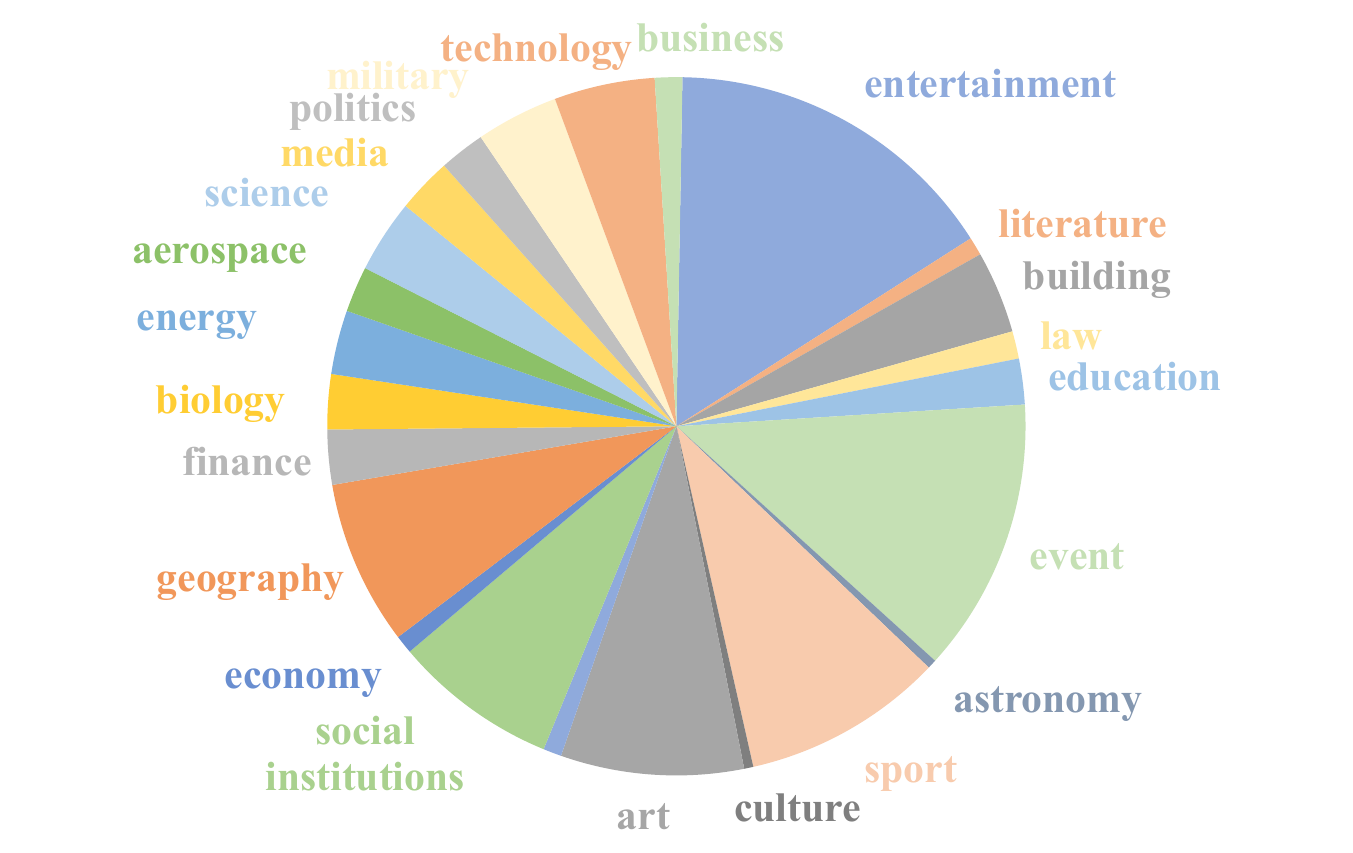}
     \caption{Domain distribution of our dataset.} 
     \label{fig:domain} 
 \end{figure}
 
Specifically, 
user instructions generally should concise (not exceeding 200 words) and directly relevant to the schema and data of the database. To simulate real-world scenarios, instructions should vary in the description style, ranging from casual chats to formal requests.
To pair with each user instruction, the annotators then retrieve a set of real texts online (such as wiki articles or news reports).
Each set contains no more than 10 documents with 3000 words.
The preferred documents are relevant to the specific databases and user instructions, and provide suitable data for extraction and population. 
The next core step is updating the databases by manually extracting values from texts according to the user instructions. The newly-added values should strictly keep to all constraints or format requirements of the databases. 
After that, the annotators map the databases and corresponding instructions to various domains, annotated according to Wikipedia's taxonomy\footnote{\url{https://en.wikipedia.org/wiki/List_of_lists_of_lists}}. This ensures that our dataset tests the generalization ability of the proposed framework across different fields and types of information.

Throughout the annotation process, we engage three Computer Science PhD students, each with research backgrounds in information extraction, to carry out the data annotation tasks. Every data instance is initially annotated by one of these annotators and subsequently reviewed by another. The annotator and reviewer discuss for any necessary adjustments until a consensus is reached on the annotation.
After finalizing the annotation, all annotators convene to assign a difficulty level—easy, medium, or hard—to each data instance. This categorization is based on multi-facet criteria, including the complexity of the database schema, the length of the source texts, and the number of values required to be populated.

\section{Tool Description}\label{sec:app_tool}
The tools can be categorized into two types:
\begin{itemize}
    \item \textbf{Information Extraction}: 
     We include tools corresponding to standard IE tasks including  (1) Named Entity Recognition, (2) Relation Extraction, (3) Attribute Extraction, and (4) Text Classification to pinpoint the entities, their attributes, relations, and categories. They set the foundation for a structured information framework necessary for database population. 
    \item \textbf{Database Integration}: 
    To materialize database updates, we employ three Database Integration Functions: 
    (5) Data Normalization then adjusts the format of the extracted information to meet database requirements, while (6) Entity Linking connects identified entities with existing entries in the database. 
    (7) Data Infilling fills in missing values by linking extracted entities with their corresponding database entries. 
    (8) Row Population involves adding new rows that adhere to the database schema and constraints. 
    (9) Column Addition introduces new columns, linking extracted entities to existing rows and populating these new columns with relevant values.
 \end{itemize}

\section{Implementation of \ours}\label{sec:app_implementation}
\begin{figure*}
     \centering
     \includegraphics[width=\textwidth]{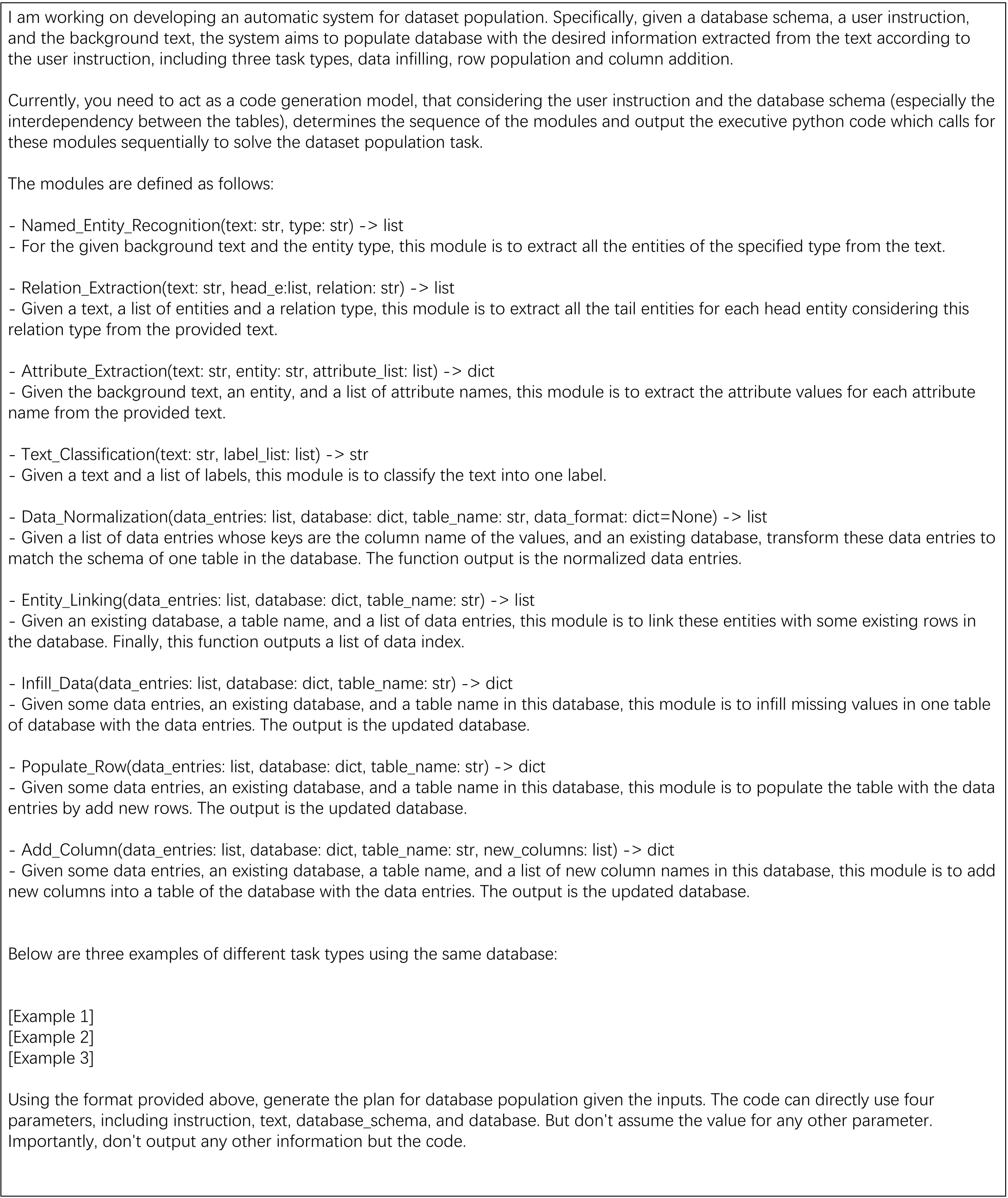}
     \caption{Prompt for the Planner.}
     \label{fig:prompt_planner}
 \end{figure*}

\begin{figure*}
     \centering
     \includegraphics[width=\textwidth]{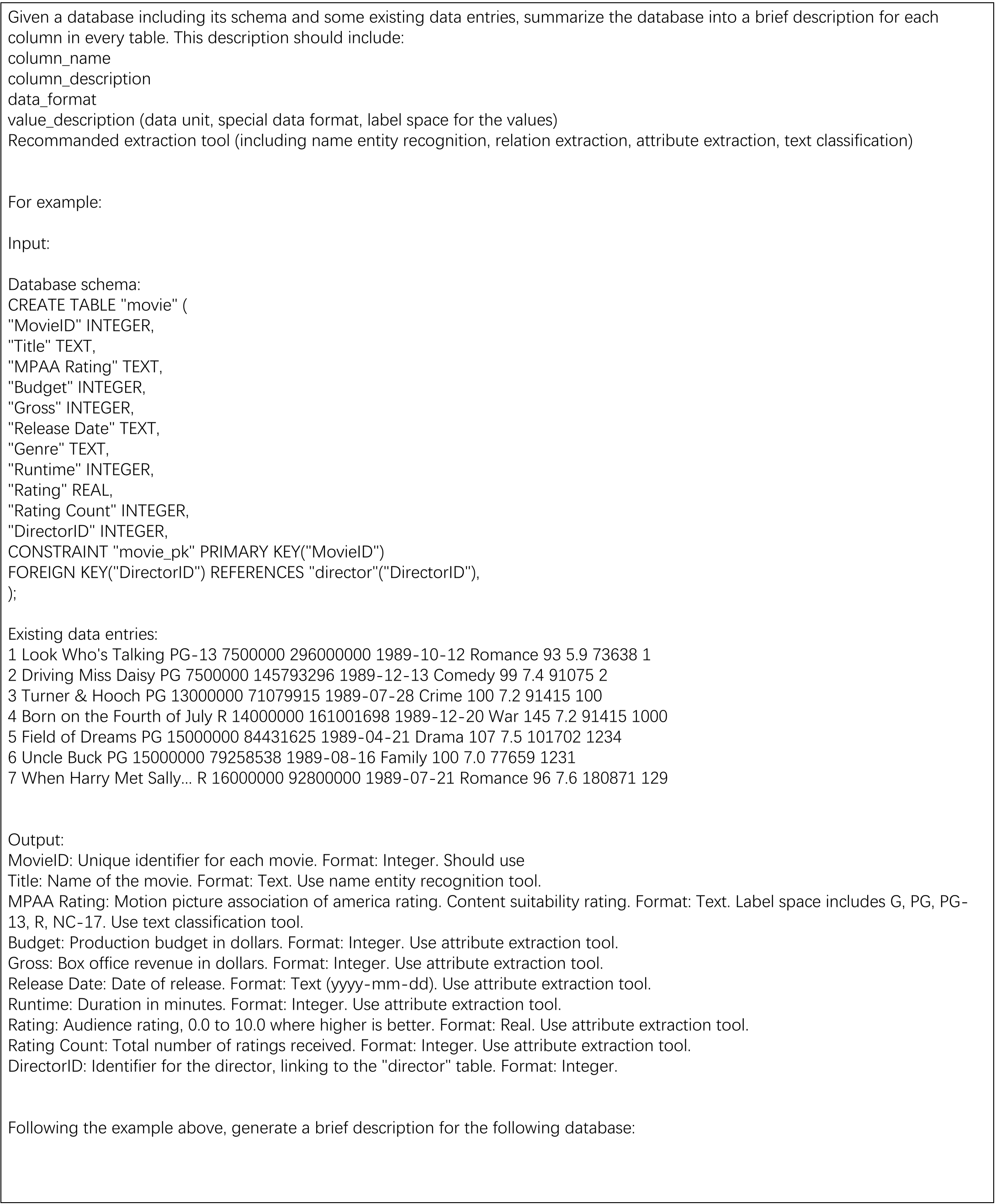}
     \caption{Prompt for the Observer.}
     \label{fig:prompt_observer}
 \end{figure*}

\begin{figure*}
     \centering
     \includegraphics[width=\textwidth]{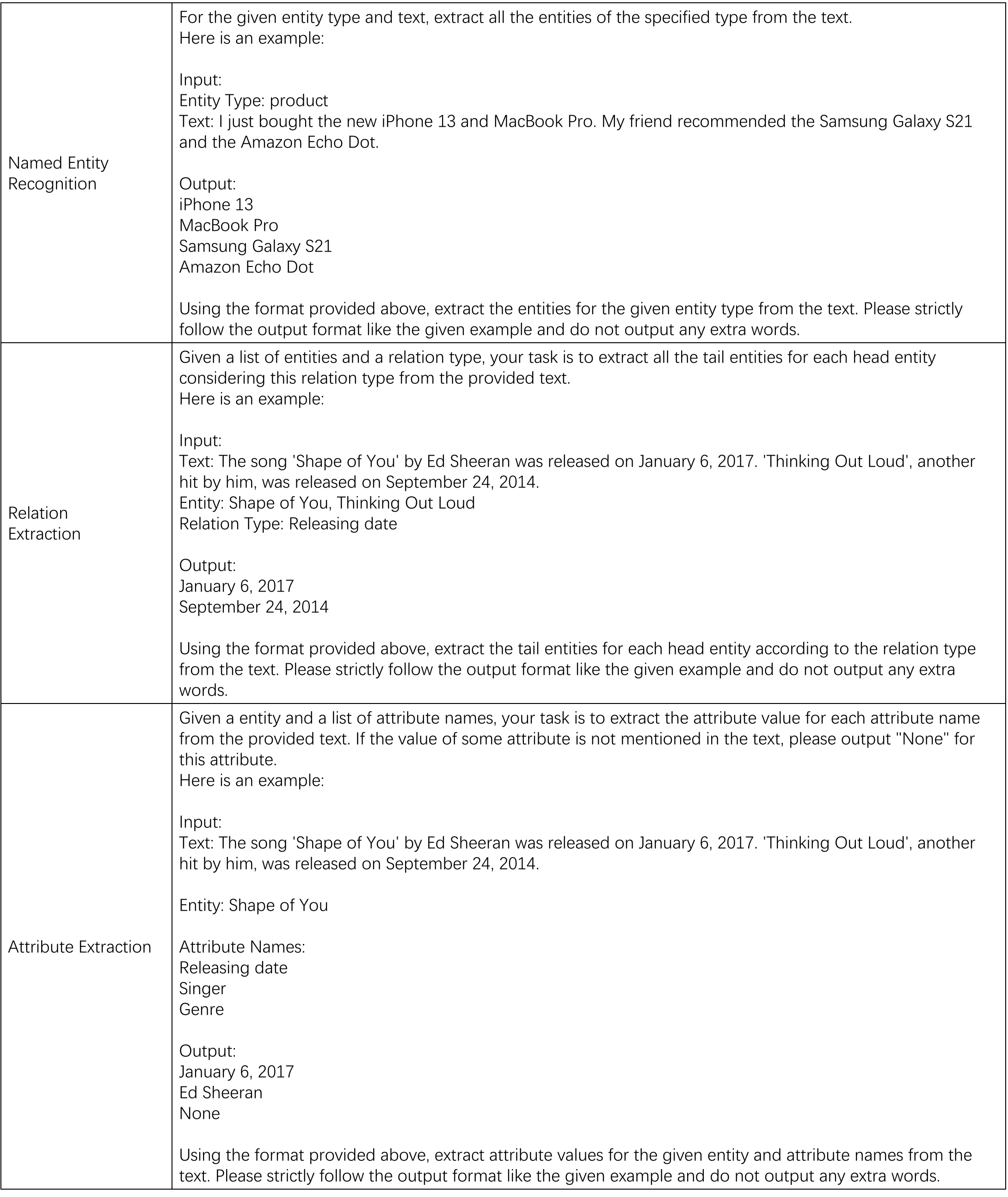}
     \caption{Prompts for the information extraction tools.}
     \label{fig:prompt_ie1}
 \end{figure*}

\begin{figure*}
     \centering
     \includegraphics[width=\textwidth]{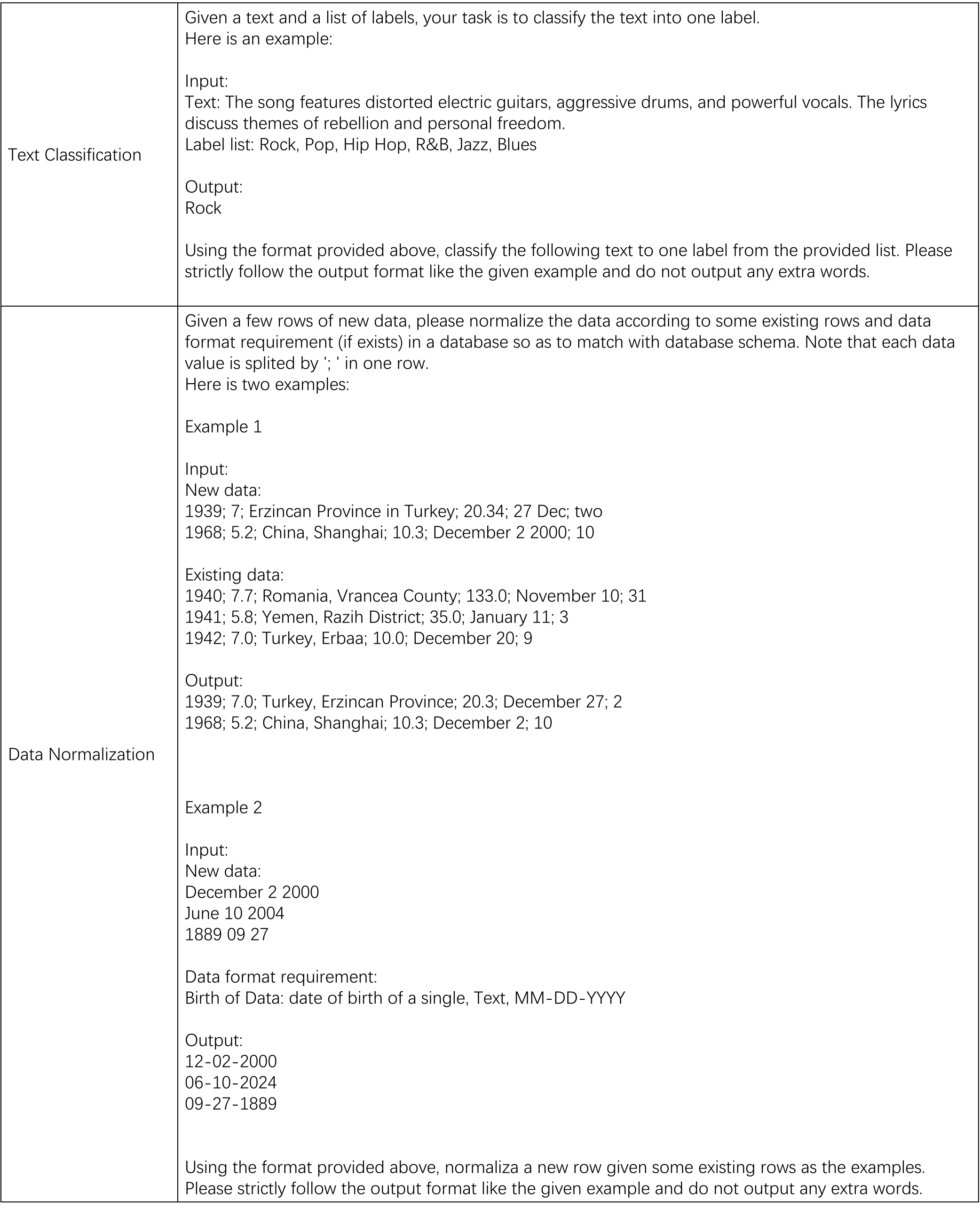}
     \caption{Prompts for the information extraction tools.}
     \label{fig:prompt_ie2}
 \end{figure*}

The planner and observer agents in our \ours\ framework are powered by the GPT4 language model \texttt{gpt-4-1106-preview}. 
The maximum number for the Planner to revise plans is 10. The whole process can restart for 2 times at most after failure. 
Specifically, the planner utilizes code from three different task types as demonstrations. The analyzer then verifies the generated code from perspectives of syntax, logic, and integrity, immediately returning any detected errors to the planner. The observer generates a summary for each table in the database to ensure the quality of observation; it selects 20 pieces of data as IE demonstrations and generate a piece of mock data for the specific task type, which may include values from multiple tables. 
For the tool library, we emulate the models with GPT for named entity recognition, relation extraction, attribute extraction, text classification and data normalization. They utilizes the demonstrations produced by the observer to achieve in-context learning. 
Table \ref{fig:prompt_planner}
- \ref{fig:prompt_ie2} shows the prompts for two agents (the planner and the observer), and four tools (named entity recognition, relation extraction, attribute extraction, text classification and data normalization).
In addition, we adopt an existing entity linking model GENRE \cite{de-cao-etal-2022-multilingual}. We transforms the extracted results into sentences using a manually designed template, to calculate the probability of linking with sentences transformed from the database's data entries.

\section{Extra Error Analysis}\label{sec:app_error}

 \begin{figure}[!h]
     \centering
     \includegraphics[width=0.5\textwidth]{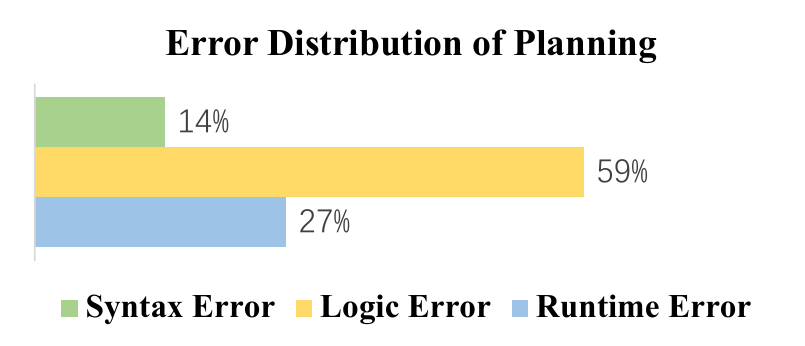}
     \caption{Error distribution of the planning step.}
     \label{fig:error2}
 \end{figure}

In the planning stage, we identified three main error subtypes: syntax errors, logic errors, and runtime errors. Syntax errors are often fixable through multiple revisions, but with more than three tables in a database, the complexity increases, and not all syntax errors can be resolved quickly. Logic errors usually stem from using incorrect tools (e.g., choosing attribute extraction over text classification for movie genres) or neglecting the interdependencies between tables (like missing foreign key relationships). Runtime errors typically occur due to tool misconfiguration (such as expecting a list instead of a string for input) or mismatches between the extracted data and the database schema (like overlooking necessary column values).

 \begin{figure}[!h]
     \centering
     \includegraphics[width=0.5\textwidth]{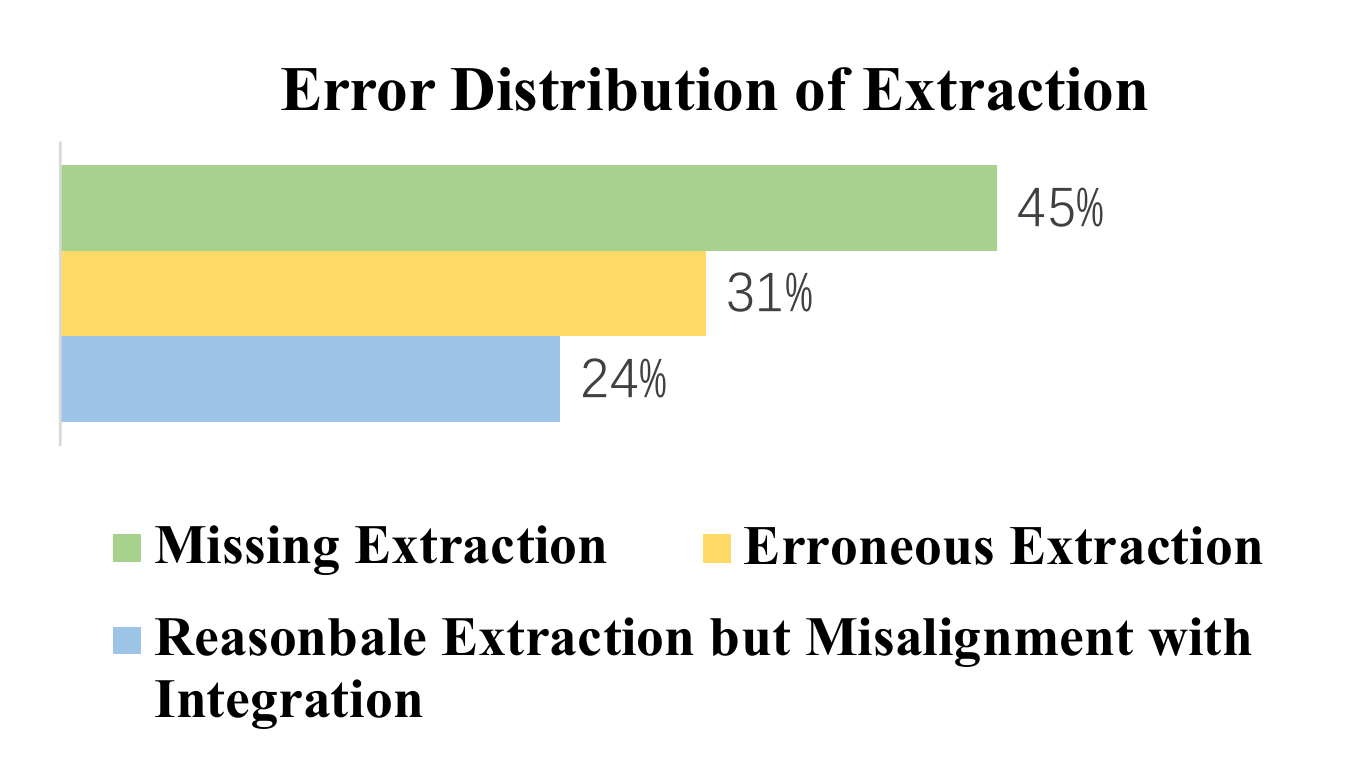}
     \caption{Error distribution of the extraction step.}
     \label{fig:error3}
 \end{figure}

In the extraction phase, the typical error subtypes are Missing Extraction, Erroneous Extraction, and Reasonable Extraction but Misaligned with Integration Requirements. Missing Extraction often occurs in long documents where IE tools may overlook sections, especially where named entity recognition models fail to capture all possible entities, leading to errors. Erroneous Extraction arises in cases with closely related information, causing confusion, such as mixing up movie premiere and release dates. The third error type involves extractions that are accurate but don't meet the specific needs for integration, like varying granularity levels in location data for an earthquake's epicenter. Demonstrations via data entry have relieved this, but there's still room for further improvement.

 \begin{figure}[!h]
     \centering
     \includegraphics[width=0.5\textwidth]{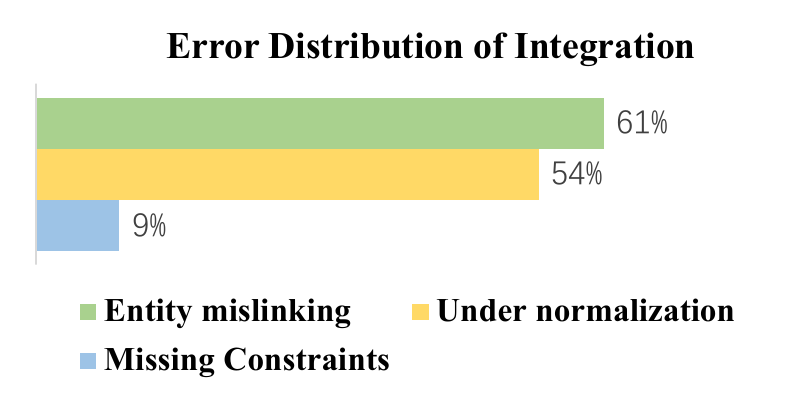}
     \caption{Error distribution of the integration step.}
     \label{fig:error4}
 \end{figure}
 
For integration, the specific error subtypes are Entity Mislinking, Under-normalization, and Data Integrity Issues. Entity Mislinking, a major bottleneck for data infilling and column addition, involves linking extracted data to the appropriate database entry, a task that becomes challenging with large databases or when database entries lack sufficient information. Under-normalization typically deals with inconsistencies in data formats, such as varying date formats, where normalization efforts might miss specific values or sacrifice efficiency for accuracy. Data Integrity Issues often arise in databases with multiple tables, where problems like missing foreign key constraints lead to the insertion of unrelated values.

\section{Case Study}
\begin{figure*}
     \centering
     \includegraphics[width=\textwidth]{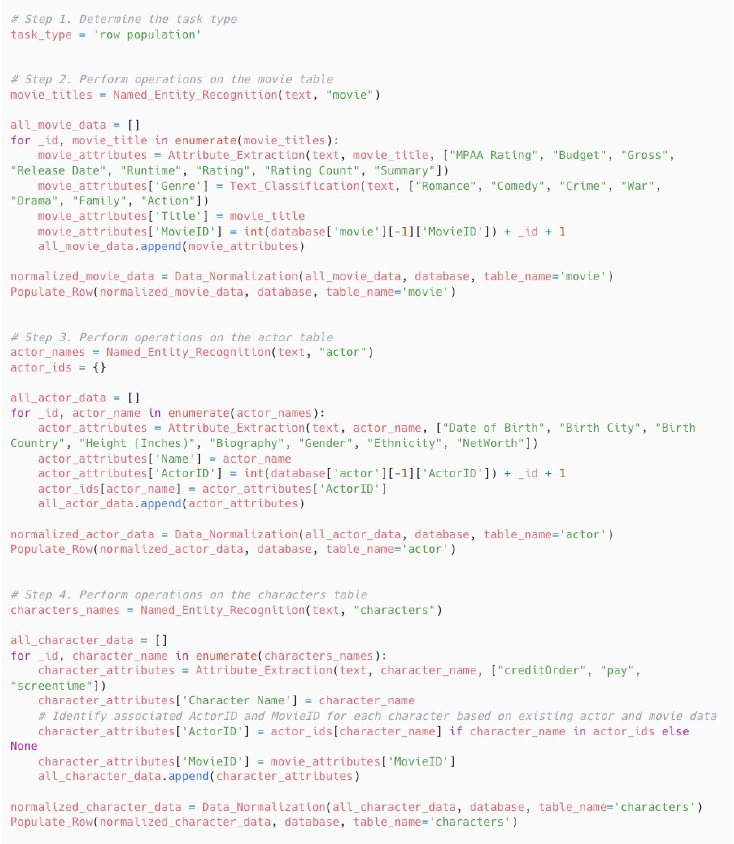}
     \caption{The complete of the generated code  for the example in Figure \ref{fig:case}. By checking the part corresponding to the \texttt{Character} table, interestingly, we observe that the model is aware of the table dependencies but the wrong key is used to find the ActorID (should be actor name instead of character name).}
     \label{fig:case_plan1}
 \end{figure*}

\begin{figure*}
     \centering
     \includegraphics[width=\textwidth]{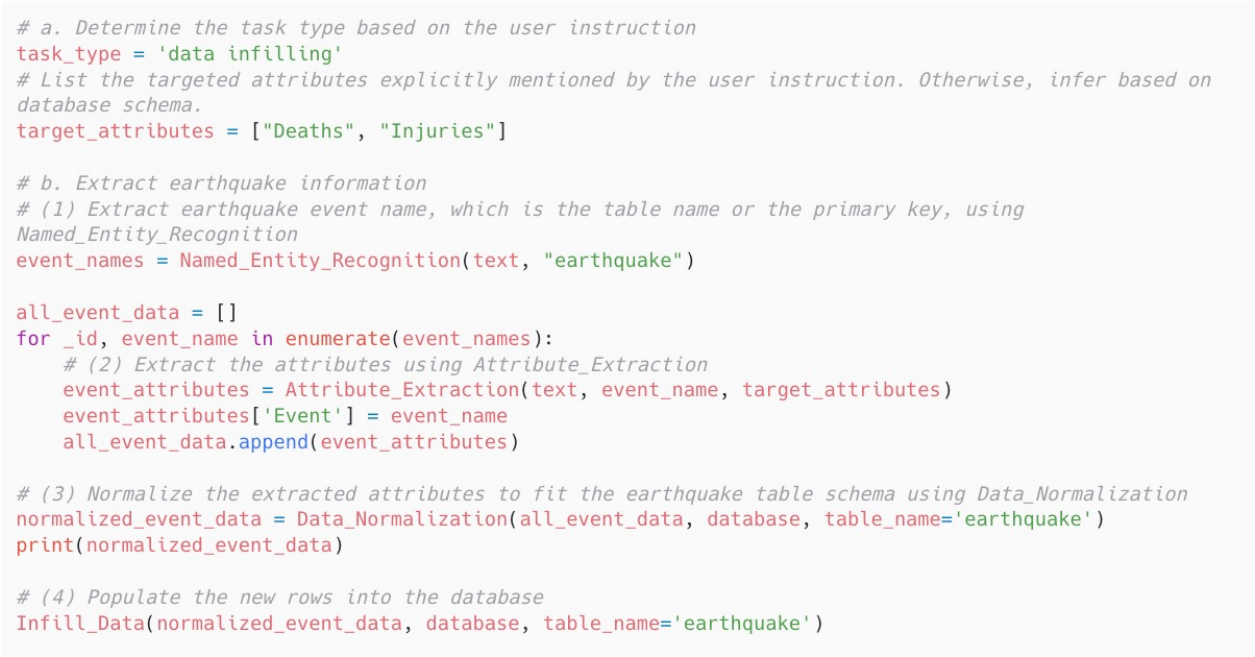}
     \caption{Example code for the data infilling task.}
     \label{fig:code_di}
 \end{figure*}

\begin{figure*}
     \centering
     \includegraphics[width=\textwidth]{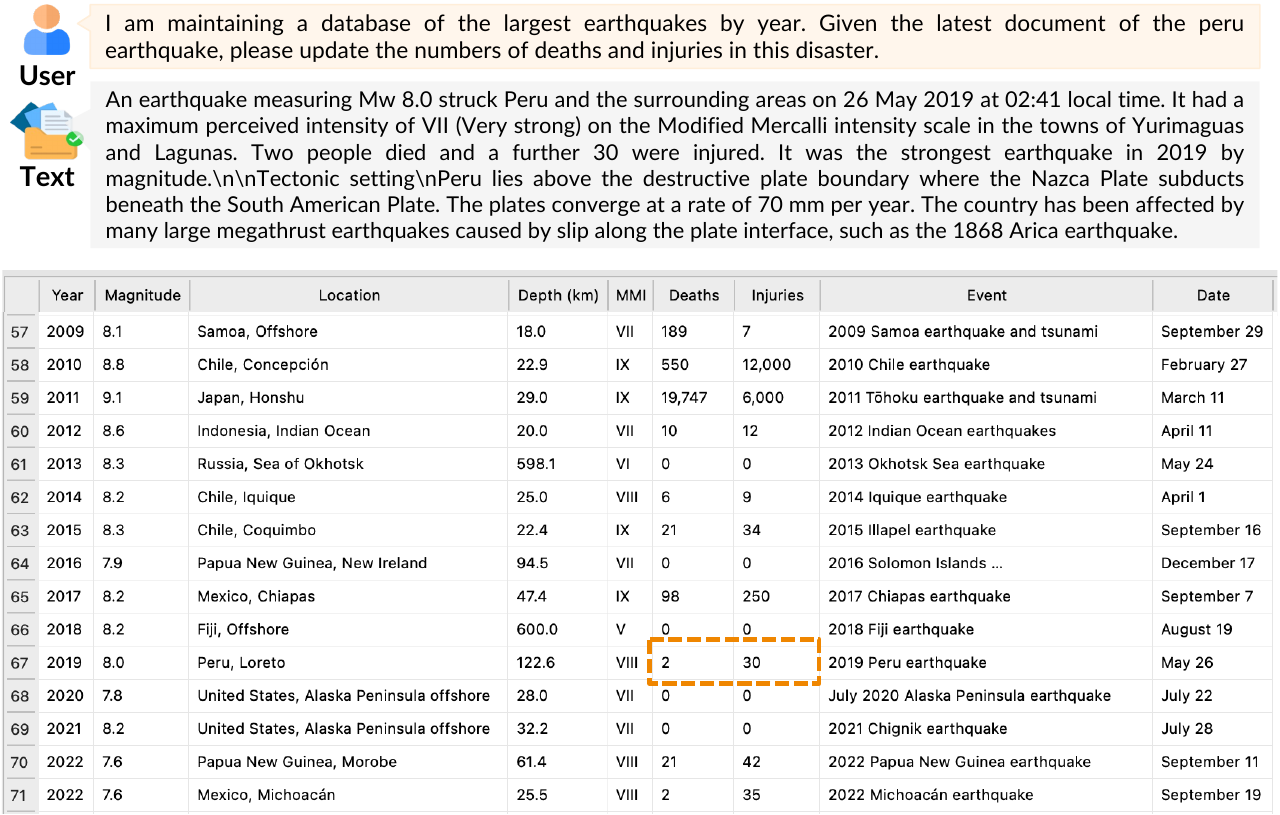}
     \caption{Case study for the data infilling task. The newly added values are framed in orange.}
     \label{fig:case_di}
 \end{figure*}

\begin{figure*}
     \centering
     \includegraphics[width=\textwidth]{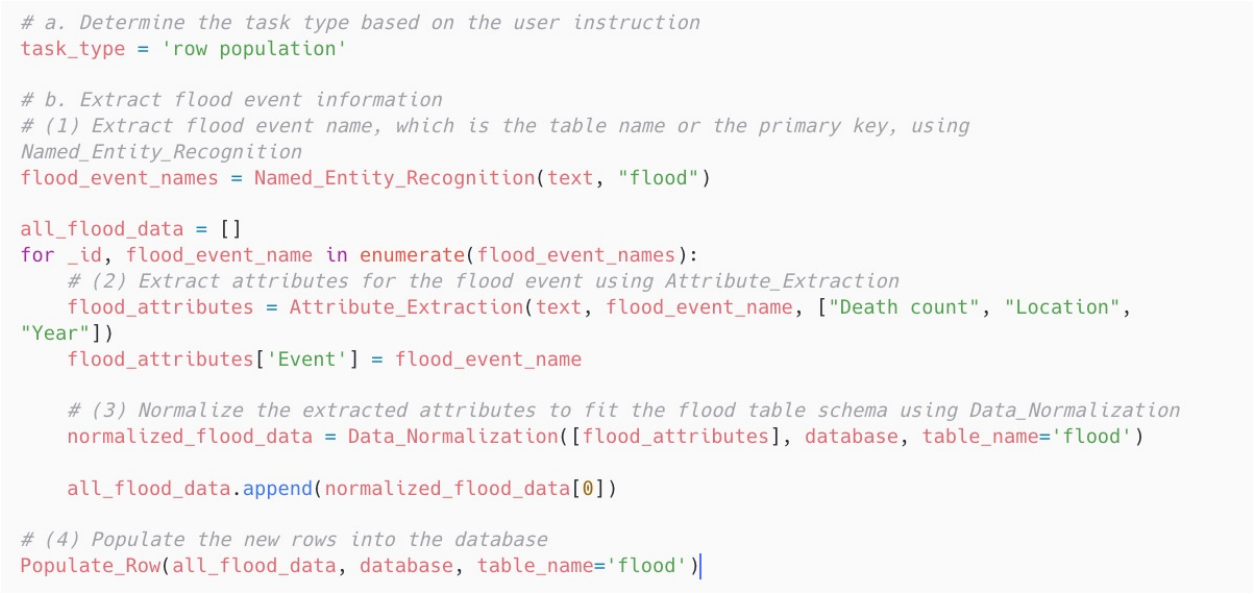}
     \caption{Example code for the row population task.}
     \label{fig:code_rp}
 \end{figure*}

\begin{figure*}
     \centering
     \includegraphics[width=\textwidth]{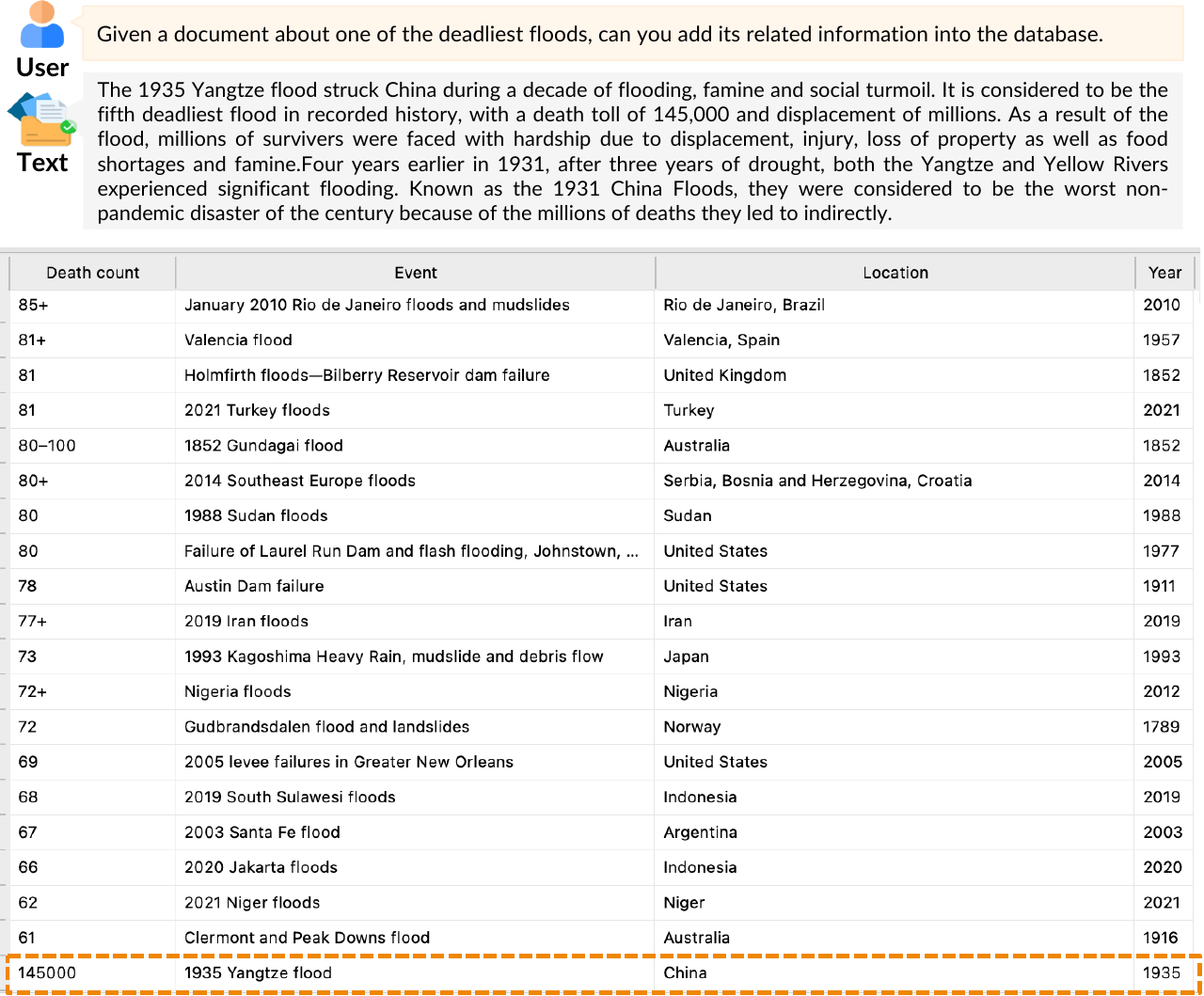}
     \caption{Case study for the row population task.  The newly added values are framed in orange.}
     \label{fig:case_rp}
 \end{figure*}

\begin{figure*}
     \centering
     \includegraphics[width=\textwidth]{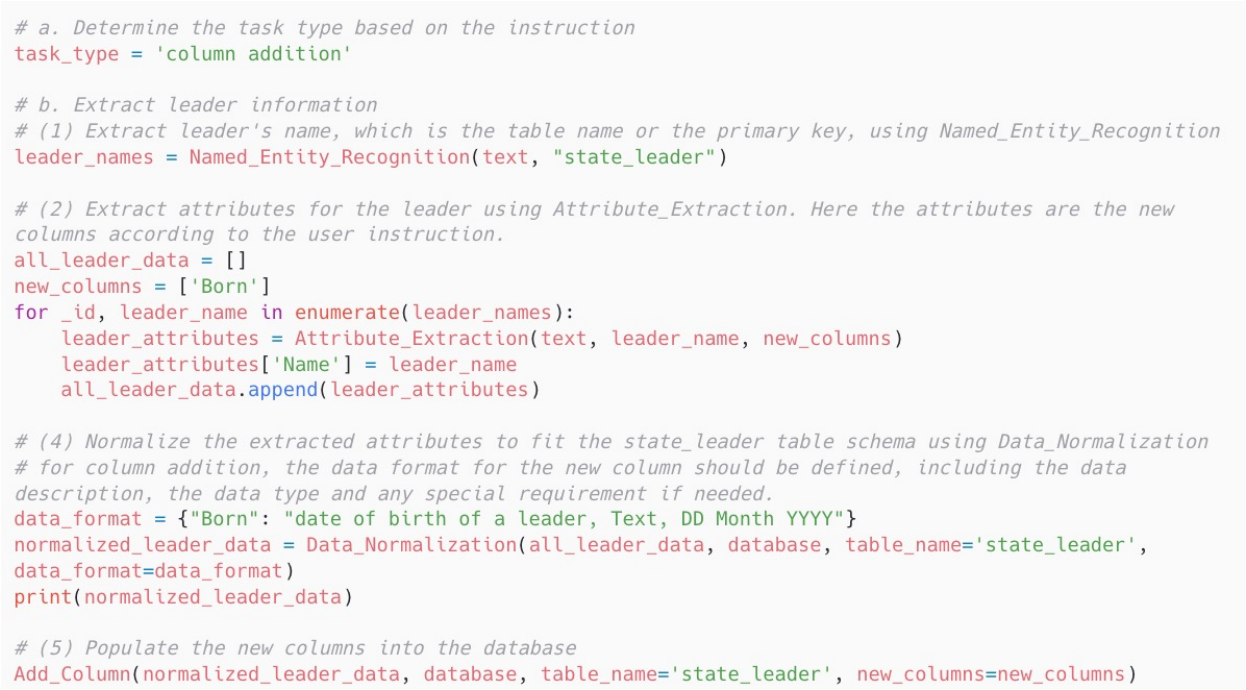}
     \caption{Example code for the column addition task.}
     \label{fig:code_ca}
 \end{figure*}

\begin{figure*}
     \centering
     \includegraphics[width=\textwidth]{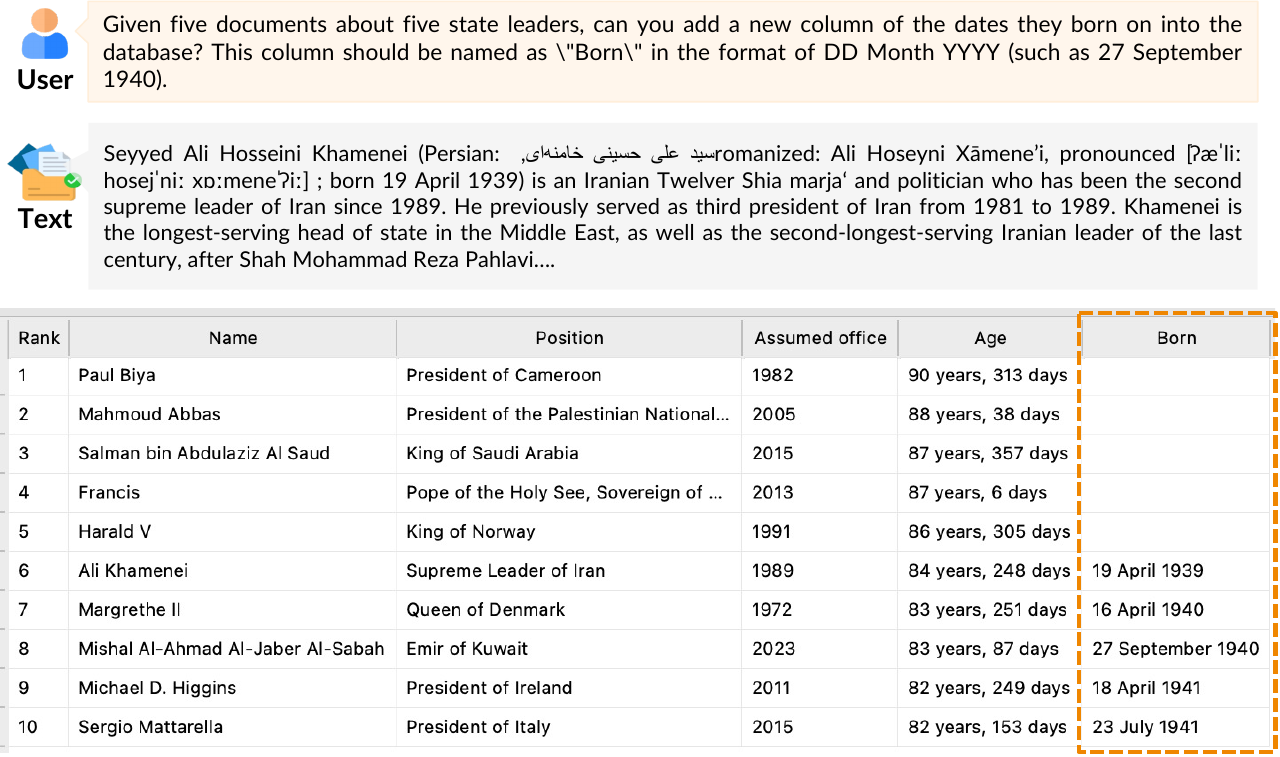}
     \caption{Case study for the column addition task.  The newly added values are framed in orange.}
     \label{fig:case_ca}
 \end{figure*}

In Figure \ref{fig:case_plan1}, we show the code for case study in Section 5.3. Meanwhile, we also present more cases for three task types, including the updated databases and the corresponding generated codes, in Figure \ref{fig:code_di}-\ref{fig:case_ca}.

 \label{sec:appendix}

\end{document}